\def\year{2022}\relax
\documentclass[letterpaper]{article} 
\usepackage{aaai22}  
\usepackage{times}  
\usepackage{helvet}  
\usepackage{courier}  
\usepackage[hyphens]{url}  
\usepackage{graphicx} 
\usepackage{natbib}  
\usepackage{caption} 
\DeclareCaptionStyle{ruled}{labelfont=normalfont,labelsep=colon,strut=off} 
\usepackage{algorithm}
\usepackage{algorithmic}

\usepackage{comment}
\usepackage{subcaption}
\usepackage{microtype}
\usepackage{enumitem}
\usepackage{amsmath}
\usepackage{xcolor}
\usepackage{multirow}
\usepackage{amsfonts}
\usepackage{booktabs}
\usepackage{bbm}
\usepackage{pifont}
\usepackage{makecell,rotating}
    \setcellgapes{5pt}

\newcommand{\footnoteref}[1]{\textsuperscript{\ref{#1}}}

\makeatletter
\def\thickhline{%
  \noalign{\ifnum0=`}\fi\hrule \@height \thickarrayrulewidth \futurelet
   \reserved@a\@xthickhline}
\def\@xthickhline{\ifx\reserved@a\thickhline
               \vskip\doublerulesep
               \vskip-\thickarrayrulewidth
             \fi
      \ifnum0=`{\fi}}
\makeatother


\newlength{\thickarrayrulewidth}
\usepackage{newfloat}
\usepackage{listings}
\floatstyle{ruled}
\newfloat{listing}{tb}{lst}{}
\floatname{listing}{Listing}
\title{NAREOR: The Narrative Reordering Problem}
\author {
    Varun Gangal \bf{\thanks{\quad Equal Contribution by the two authors}} \textsuperscript{\rm 1}
    Steven Y. Feng \footnotemark[1] \textsuperscript{\rm 1}
    Malihe Alikhani \textsuperscript{\rm 2}
    Teruko Mitamura \textsuperscript{\rm 1}
    Eduard Hovy \textsuperscript{\rm 1}
}
\affiliations {
    \textsuperscript{\rm 1} Language Technologies Institute, Carnegie Mellon University\\
    {\tt \{vgangal,syfeng,teruko,hovy\}@cs.cmu.edu } \\
    \textsuperscript{\rm 2} School of Computing and Information, University of Pittsburgh\\
   {\tt malihe@pitt.edu }
}

\begin{document}

\maketitle

\begin{abstract}
Many implicit inferences exist in text depending on how it is structured that can critically impact the text's interpretation and meaning. 
One such structural aspect present in text with chronology is the order of its presentation. For narratives or stories, this is known as the \textit{narrative order}. 
Reordering a narrative can impact the temporal, causal, event-based, and other inferences readers draw from it, which in turn can have strong effects both on its interpretation and interestingness. In this paper, we propose and investigate the task of \textit{Narrative Reordering} (NAREOR) which involves rewriting a given story in a different narrative order while preserving its plot. 
We present a dataset, NAREORC, with  
human rewritings of stories within ROCStories in non-linear orders, and conduct a detailed analysis of it. Further, we propose novel task-specific training methods with suitable evaluation metrics. We perform experiments on NAREORC using state-of-the-art models such as BART and T5 and conduct extensive automatic and human evaluations. We demonstrate that although our models can perform decently, NAREOR is a challenging task with potential for further exploration. 
We also investigate two applications of NAREOR: generation of more interesting variations of stories and serving as adversarial sets for temporal/event-related tasks, besides discussing other prospective ones, such as for pedagogical setups related to language skills like essay writing and applications to medicine involving clinical narratives.


%
\end{abstract}

\section{Introduction}
\label{sec:intro}
From the onset of language, storytelling has been crucial to the transmission of knowledge 
\cite{ramanujan1991three}. It has been well-established that readers remember only an abstract representation of stories \cite{schank1972conceptual}.
Before the printing press, classes engaged with oral teaching of scriptures, such as rabbis, underwent extensive training to reproduce them with no distortion \cite{bos1995jewish}. 
Formally analyzing story structure commenced with the ancients, through works like Aristotle's \textit{Poetics} \cite{halliwell1998aristotle}. 
These studies led to the concept of a \emph{narrative}, distinct from story events.

\begin{figure}
\begin{tabular}{@{}ll@{}}
\includegraphics[width=0.46\textwidth]{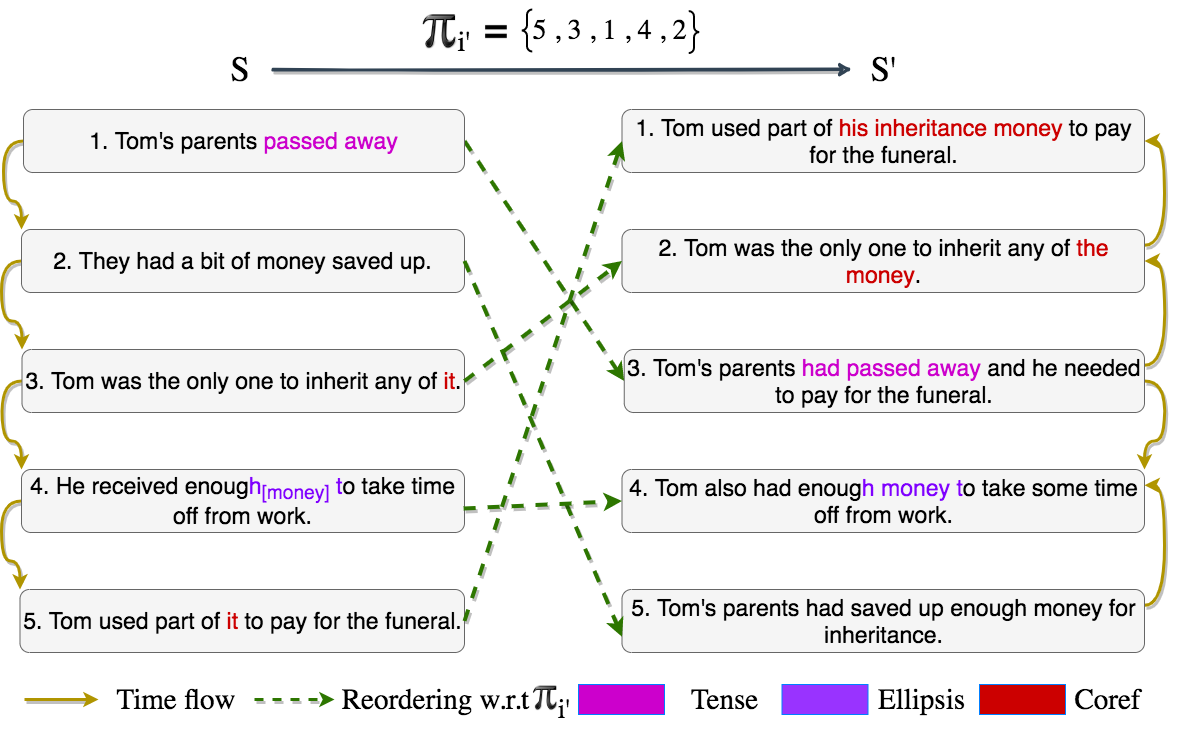}\\
\end{tabular}
  \caption{\small \label{tab:qualitative_1} Example of our task and dataset, with original input story $S$ on the left, target narrative order $\pi_{i'}$ on the top, and human rewritten story $S'$ on the right.}
\end{figure} 


For a story, there are two \textit{orders}: the chronological order of events as they happened and their order as presented in text. These have been analyzed under different names \cite{propp2010morphology}. 
We refer to them as \textit{story order} and \textit{narrative order}, or \textit{story} and \textit{narrative}, respectively. 
\citet{genette1983narrative} enlists typical orders observed in writing. 
A \textit{linear} order narrates events in same sequence as story order. The \textit{in medias res} order starts with events in the middle, goes back to the start, then proceeds to the end. Changing from near-\textit{linear} to more ``interesting" 
orders is prevalent in cinema, e.g. \textit{The Imitation Game} starts with Turing's post-WWII 1951 interrogation. \textit{Memento} and \textit{Naked Lunch} are known for their esoteric narrative orders - loosely described as retrogade (reverse of linear) and syllepsis (lacking chronological logic), respectively. 

\citet{morgan2017narrative} explains how narratives surpass ``mere chronicle". Narrative orders of presenting materials in scientific \emph{explanations} directly affects how 
researchers \emph{interpret} 
and \emph{understand} them since the order implies not only temporal but other inferences about causality, processes of change, etc. Narrative order can thus influence \emph{model explainability}, especially for explanation generation \cite{rajani-etal-2019-explain}, a recent area-of-interest \cite{wiegreffe2021teach}.

In this work, we do not delve into the complex and somewhat subjective question of \emph{which} narrative order is most suitable or ``interesting". We focus on \emph{how} a given story in \emph{linear} narrative order can be rendered in a specified, \emph{non-linear}, \emph{target} order while preserving plot.\footnote{Code+data at \url{github.com/vgtomahawk/NAREORCamReady}} 
We call this \textit{Narrative Reordering}, or NAREOR. To the best of our knowledge, we are the first to propose and investigate this task.

Our work is not entirely adrift from past research in this vein. \citet{montfort2007ordering} tries generating fiction narratives from basic existent-event info with a special focus on narrative order, using a rule and planning approach. Unlike our work, their rule-based system does not involve learning. Moreover, being generation in a given narrative order from unstructured story elements rather than reordering an existing story, their setting does not require solving challenges such as disentangling events from stories which are inherent in NAREOR.

Formally, NAREOR involves reordering a story $S$ with sentences $s_{1},s_{2},...,s_{n}$ to a reordered story $S'$ with sentences $s'_{1},s'_{2},...,s'_{n}$ according to a given target narrative order $\pi_{i'}$. $\pi_{i'}$ is 
a permutation $\{\pi_{i'}|\pi_{i'}: i' \rightarrow f(i'); 1 \leq i' \leq n; f(i')=i\}$ mapping from target sentence\footnote{For simplicity, we assume \emph{narrative} to break up into sentence units. Our task is still very challenging as shown through this paper.} indices $i'$ to original sentence indices $i$, where $f$ is a one-to-one and onto function from $\{1,2\ldots n\}$ to itself. In practice, we write $\pi_{i'}$ as the sequence $\{i=f(i')\}_{i'=1}^{i'=n}$ ($f$ and $i'$ become implied).




NAREOR's challenges are evident from the example in Figure~\ref{tab:qualitative_1}. Simply reordering sentences is far from sufficient, as rewritten text must be adjusted 
to handle coreference, tense, and other discourse dependencies. For example, narrative order affects tense since it can change the first 2 of 3 Reichenbach times \cite{reichenbach1947elements} that together determine tense - speech, reference, and event time. NAREOR involves pinpointed and critical edits; a single missed or incorrect edit can result in an entirely different or invalid plot. 
Since $\pi_{i'}$ can be seen as a \emph{control}, NAREOR is a controllable generation task (see Appendix A for discussion). 

NAREOR is also a novel form of story-level paraphrasing and can be used to generate more interesting variations of stories (\S\ref{sec:interest_results}). Outputs can also serve as challenge sets for temporal or event-based tasks such as sentence ordering to assess the temporal reasoning capabilities of models (\S\ref{sec:applications}). NAREOR can also be potentially useful for pedagogical setups related to language skills such as essay writing, and applications to medicine involving clinical narratives (\S\ref{sec:applications}). 

To complement NAREOR, we present a dataset, NAREORC, with human rewritings of stories from ROCStories \cite{mostafazadeh2016corpus} in non-linear orders. We conduct a thorough analysis, examining various ways humans modify the text when reordering (\S\ref{sec:dataset}). We perform experiments with BART, T5, and GPT-2 on NAREORC using novel, task-motivated training methods we propose (\S\ref{sec:methodology}). We evaluate our models with both an automatic and human evaluation along with qualitative analysis (\S\ref{sec:results_and_analysis}). We demonstrate that our proposed training methods are effective but have room for further improvement. We illustrate that NAREOR is indeed a challenging task with potential for further exploration.

\section{Dataset: NAREORC}
\label{sec:dataset}
\subsection{Dataset Construction}
\paragraph{Source Corpus:} ROCStories has $\approx$ 98.5K five-sentence English stories. For the dev and test splits, each example contains a four-sentence story prefix with a one-sentence coherent and incoherent ending. We treat the coherent endings as the fifth sentences for NAREORC's dev and test stories.

\paragraph{Assigning Target Narrative Orders:} The target narrative order $\pi_{i'}$ is not part of the ROCStories input. 
We devise a randomized procedure to assign a reasonable $\pi_{i'}$ for each example. We sample 3 permutations from the set of non-identity n!-1 permutations.\footnote{In our case, $n=5$ as we experiment with ROCStories.} We find Kendall $\tau$ correlations 
\cite{kendall1938new} between identity permutation $I_{n}$, \{1,2,3,4,5\}, and each of the three permutations, retaining the lowest as $\pi_{i'}$. We prefer this to sampling 
at random because we want our examples to be sufficiently non-trivial w.r.t. the task.

\paragraph{Supervised \& Unsupervised Splits:} We set aside 600, 200, 200 stories from train, dev, and test splits of ROCStories. These act as NAREORC's trainSup, devSup, and testSup splits, for which we collect human references. Remaining stories in each ROCStories split are retained as trainUnsup, devUnsup, and testUnsup of size 95161, 1671, 1671.

\paragraph{Human Annotation:} For trainSup and devSup, we annotate one reference per example. For testSup, we collect two each to help reference-based metrics. We conduct our study on AMT. To understand task difficulty, we ask a ``Hardness" question with options \textit{VeryEasy}, \textit{Easy}, \textit{Moderate}, \textit{Hard}, \textit{VeryHard}. On average, annotators found $\approx$70\% of rewritings to be \textit{Moderate} or \textit{Hard}, demonstrating that NAREOR is quite difficult even for humans. More details in Appendix B.

\subsection{Dataset Analysis}
\label{sec:dataset_analysis}

\paragraph{Overall Statistics}\mbox{}\\
We find human-rewritten stories $S'$ are $\approx$1.2x as long as input stories $S$ on avg in words and characters. We expect this given the narrative reordered story favors resolution of sentence-order dependent elements like ellipses ($s_{4}$ and $s'_{4}$ in Figure \ref{tab:qualitative_1}) and pronouns ($s_{3}$ and $s'_{2}$ in Figure \ref{tab:qualitative_1}) to explicit forms. It also requires insertion of time expressions (e.g \textit{Before that,} 3rd row, Table \ref{tab:categoriesTwo}) to clarify the now disrupted flow.

Unique n-gram ratio $\text{UR}_{n}(S)$ is the fraction of unique n-grams of length $n$ in $S$. We observe all three mean URs ($n=1,2,3$) to decrease from input to reference story. $\text{UR}_{1}$: 0.692$\rightarrow$0.669, $\text{UR}_{2}$: 0.940$\rightarrow$0.931, $\text{UR}_{3}$: 0.989$\rightarrow$0.984. Increased n-gram repetition could have reasons similar to length increase, 
causing cross-sentence repetition. 
Figure \ref{tab:qualitative_1} demonstrates this: $S$ only has one instance of \textit{money}. Conversion of \textit{inherit any of it} ($s_{3}$) $\rightarrow$ \textit{inherit any of the money} ($s'_{2}$) and 
\textit{enough to take time} ($s_{4}$) $\rightarrow$ \textit{enough money to take some time} ($s'_{4}$), among other changes, results in four in S'.

\paragraph{How Verb Forms Change}\label{subsec:howVerbsChange}\mbox{}\\
We note changes in occurrence distribution 
across verb-related \textit{pos} tags from $S$ to $S'$ using NLTK's \textit{pos} tagger. Gerund fraction (\textit{pos}=VBG) (e.g. \textit{I like \textbf{playing}}) increases 7.7\%$\rightarrow$9.5\%. Past participle fraction (\textit{pos}=VBN) (e.g. \textit{He had \textbf{broken} it}) $\approx$ doubles, 6.5\%$\rightarrow$12.4\%. Past tense fraction (\textit{pos}=VBD) (e.g. \textit{He \textbf{broke} it}) decreases 60.9\%$\rightarrow$54.6\%. Other verb-related \textit{pos} fractions remain fairly constant. Increase in past participle can be explained by frequent conversion to past perfect tense during reordering (e.g. \textit{parents \textbf{passed} away}$\rightarrow$\textit{parents \textbf{had passed} away} in Figure \ref{tab:qualitative_1}).

\paragraph{How Narrative Reordering Alters Sentences}\label{sec:reordering_analysis}\mbox{}\\
We look at corresponding sentence pairs $\{s_{i},s'_{i'}\}$ in each story, 
specifically 4 linguistic change types - ellipsis, tense, time expressions (timexes), coreference. We tried detecting these using off-the-shelf tools, and did not find any for ellipsis. Timex detectors like SUTime \cite{chang2012sutime} only mark strict timexes (e.g. \textit{last Sunday}) but not others (e.g. \textit{before midsems}). We hence hand-annotate these four for each $\{s_{i},
s'_{i'}\}$ per testSup example. These are further described in Table \ref{tab:categoriesTwo}. We find over half (51.5\%) the examples show $\geq$3 of 4 change types at once, and 89.5\% show $\geq$2. This shows that NAREOR requires performing different changes in tandem.

\begin{table*}[!ht]
    \small
    \setlength{\tabcolsep}{2pt}
    \begin{tabular}{l l}
    \toprule 
          \begin{tabular}{l} Change Type \end{tabular}  &   \begin{tabular}{l} Story Examples with Changes Highlighted \end{tabular}    \\ \midrule
        
        \begin{tabular}{l} {Ellipsis} \\ (Sent: 5.7\%) \\ (Stor: 27.5\%) \end{tabular}  & \begin{tabular}{p{0.85\textwidth}} \textbf{S}: 1. All of the Ross family has red hair, except Henry. 2. Henry has blonde hair that is very curly. 3. Henry's father often teases Henry's mother about the mailman. 4. The mailman has blonde, curly hair, but he is very ugly. \textit{5. His dad's teasing makes Henry feel bad.} ; $\bf{\pi_{i'}}$: $\{1, 5, 4, 2, 3\}$ \\

        \textbf{S'}: 1. All of the Ross family has red hair, except Henry. \textit{2. His dad's teasing \textbf{about the mailman}} \textit{makes Henry feel very bad.} 3. This is because the mailman has blonde, curly hair, but he is very ugly. 4. Henry also has blonde hair that is very curly. 5. Henry's father often teases Henry's mother about the mailman.
         \end{tabular}   \\ 
         \midrule 
        \begin{tabular}{l} {Tense} \\ (Sent: 19.1\%) \\ (Stor: 64.0\%) \end{tabular} & \begin{tabular}{p{0.85\textwidth}} \textbf{S}: 1. Sam bought a new SUV. 2. It was all wheel drive. 3. He figured he would take it off road. \textit{4. He} \textit{\textbf{hit} a few hard bumps and broke his suspension.} 5. Sheepishly, he brought it to the dealership for repair. ; $\bf{\pi_{i'}}$: $\{2, 3, 5, 1, 4\}$

        \textbf{S'}: 1. Sam's SUV was an all wheel drive. 2. He thought he could take it for a spin off road. 3. Embarrassed by the outcome of his drive, Sam took the car to the dealership for repair. 4. He had just bought the SUV. \textit{5. The car \textbf{had hit} a few hard bumps and the suspension broke when Sam took it off road.}
        \end{tabular}   
        \\ \midrule 
        
        \begin{tabular}{l} {Timexes} \\ (Sent: 34.0\%) \\ (Stor: 85.5\%) \end{tabular}   & \begin{tabular}{p{0.85\textwidth}} \textbf{S}: 1. There was once a kitten that did not have a home. \textit{2. The poor kitten walked around cold and} \textit{hungry.} 3. One day, a nice lady let the kitten into her home. 4. The woman gave the kitten food and a bed. 5. The kitten was happy to be adopted. ; $\bf{\pi_{i'}}$: $\{4,2,5,1,3\}$ \\

        \textbf{S'}: 1. A woman gave a home to a cat. \textit{2. \textbf{Before that} it was cold and hungry}. 3. It made the cat happy to have a home. 4. The little cat originally was homeless. 5. But in the end, it met the nice woman and she let it in.
        \end{tabular}  \\ 
        \midrule 
        \begin{tabular}{l} {Coreference} \\ (Sent: 20.7\%) \\ (Stor: 71.5\%) \end{tabular}   & \begin{tabular}{p{0.85\textwidth}}
        \textbf{S}: 1. Jimmy wandered around the city looking for a place for a soda. 2. Before he knew it, he was in an unfamiliar area. 3. He was scared of strangers and didn't want to ask anyone. 4. Soon a policeman came by and asked if he was lost. \textit{5. \textbf{He} told him that he was lost.} ; $\bf{\pi_{i'}}$: $\{5,4,2,1,3\}$ \\

        \textbf{S'}: \textit{1. \textbf{Jimmy} told a police officer that he was lost.} 2. He was lucky the police showed up in the first place. 3. He had no idea where he was. 4. He had wandered off when trying to find somewhere to buy a soda. 5. It was pretty terrifying being all alone in a mysterious area with strangers.
        \\ \end{tabular}   \\ \bottomrule 
    \end{tabular}
    \caption{\small Sentence pairs in testSup stories are annotated for 4 linguistic change types common in NAREORC. Sent denotes \% of sentence pairs showing that change type. Stor denotes story pairs $(S,S')$ where $\ge$ one sentence pair shows that change type.}
    \label{tab:categoriesTwo}
\end{table*}



\section{Methodology}
\label{sec:methodology}
\subsection{Training Methods}
\label{sec:models}

We introduce two task-specific training methods.


\paragraph{NAR-denoise (NAR-d)}\mbox{}\\
\label{sec:NAR-denoise}
This is partially inspired by how humans rewrite; a common approach is to first reorder sentences naively (simply swap positions), then make other changes. NAR-d attempts to mimic this, learning to convert from naive orderings to high-quality text. It involves two stages of model training.
\begin{enumerate}[wide=0pt,noitemsep,topsep=2pt]
    \item \textbf{Denoise-1S:} Stage 1 is unsupervised training through story-level denoising. We use trainUnsup without human-written reorderings, and simulate them using the original human-written ROCStories (the outputs during training). Deletion and swapping of tokens are used to create inputs from these stories that simulate naive reorderings. This noising aims to emulate the reverse of the content editing that occurs during NAREOR. 
    Specifically, we randomly delete 12.5\% of tokens and swap another 12.5\%. We found human-rewritten stories were, on average, in combination of token length (longer) and swappings, $\approx$25\% different from the originals. We split this 
    between deletion and swapping to approximate naively-reordered stories. 
    Story sentences $S$ are first reordered as per $\pi_{i'}$ to produce $S'_{naive}$, then each is edited to fit the new narrative. 
    We swap tokens as humans often swap words like coreferent mentions based on how the narrative order changes. 
    Hence, this stage learns to denoise text by converting noised versions to human-written text.
    \item \textbf{Denoise-2S:} The second stage is supervised training atop the model above. The inputs are the 600 original stories in trainSup, with sentences naively reordered as per target narrative order $\pi_{i'}$ to $S'_{naive}$, and the outputs are the human rewritings of these. The model learns to further translate from naively-reordered text to fluent human-written text.
\end{enumerate}

\paragraph{NAR-reorder (NAR-r)}\mbox{}\\
\label{sec:NAR-reorder}
Unlike NAR-d, NAR-r models themselves handle reordering given the target order rather than naive reordering beforehand.
\begin{itemize}[wide=0pt,noitemsep,topsep=1pt]
\item \textbf{Input Encoding Scheme:} We describe how the task input $\{$S,$\pi_{i'}\}$ is encoded as a token sequence for both Stage-1 and 2 training. To enable the model to distinguish different sentences, we prefix each s $\in$ S with a tag from {$<$a$>$} to {$<$e$>$}. We specify $\pi_{i'}$ as a sequence of these, separated from S by {$<$sep$>$}. NAREOR involves rearranging mention types among coreference chains (see \S\ref{sec:reordering_analysis}), so we use NeuralCoref \cite{neuralcoref} to detect these chains. For each, we assign a unique uppercase tag ({$<$X$>$}) to replace its mentions. At the end of the input, we list each tag and the {head mention} of its coreference chain in order. We then append {$<$st$>$} to mark the end of the input. 
An illustration of the scheme follows: 
{\small \textit{{$<$a$>$} Since I had front seat tickets, I was able to directly see {$<$X1$>$}. {$<$b$>$} {$<$X1$>$} tried to reach out with {$<$X1$>$ $<$X2$>$}. {$<$c$>$} I grabbed {$<$X2$>$} and {$<$X1$>$} pulled me on stage. {$<$d$>$} {$<$X1$>$} began to sing. {$<$e$>$} The concert had started. {$<$sep$>$} {$<$e$>$ $<$d$>$ $<$a$>$ $<$b$>$ $<$c$>$} {$<$X1$>$} {The music artist} {$<$X2$>$} {her hand} {$<$st$>$}}\normalsize}{}

\item \textbf{Reorder-1S:} We use examples from trainUnsup for stage 1. It is problematic to train for the forward direction of our task $S,\pi_{i'} \rightarrow S'$ since $S'$ is not known. Approximating $S'$ using $S'_{naive}$ 
would hurt output fluency. We instead train in the inverse direction $S'_{naive},\pi^{-1}_{i'} \rightarrow S$, where $\pi^{-1}_{i'};\pi^{-1}_{i'}(\pi_{i'})=I_{n}$ is the inverse permutation of $\pi_{i'}$. 
To reduce train-test mismatch, 
we use the inverse formulation half the time, and an autoencoding one, i.e. $S,I_{n} \rightarrow S$ the other half. 
\item \textbf{Reorder-2S:} trainSup examples are used to further finetune on reorder-1S. We train in the task direction $S,\pi_{i'} \rightarrow S'$.
\end{itemize}

\subsection{Chosen Models}\label{sec:chosen_models}
We choose several pretrained generation models: GPT-2, BART, and T5. We finetune all using both our training methods to produce denoise-1S (d-1S), denoise-2S (d-2S), reorder-1S (r-1S), and reorder-2S (r-2S) versions. GPT-2 \cite{radford2019language} is a Transformer-based language model trained on \textit{WebText}. 
BART \cite{lewis-etal-2020-bart} and T5 \cite{JMLR:v21:20-074} are Transformer seq2seq models. BART is trained as a denoising autoencoder to reconstruct original from noised text. T5 
is designed to be effective for transfer learning. We use HuggingFace's implementations of their base versions.\footnote{See \S\ref{sec:training_finetuning} for further training/finetuning details.}

\subsection{Automatic Evaluation Metrics}
\label{sec:metrics}



\paragraph{Reference-Based Metrics} assess the similarity between generated text and human-written references. We use \textbf{BLEU} \cite{papineni-etal-2002-bleu}, \textbf{METEOR} \cite{banerjee2005meteor}, and \textbf{BERTScore} \cite{zhang2019bertscore}. 
We compare generated text with the two 
references per testSup example.\footnote{Correlates well with human evaluation as shown in \S\ref{sec:results_and_analysis}.}

\paragraph{Target Order Fidelity (TOF)}
is defined as how closely the reordered text matches the given target narrative order. E.g. given $S=\{s_{1},s_{2},s_{3}\}$, $\pi_{i'} = \{3,2,1\}$, and $S'=\{s'_{1},s'_{2},s'_{3}\}$, we wish to see if $s_{1}$ has correctly been translated to $s'_{3}$. We introduce \textbf{TOF-METEOR} and \textbf{TOF-BERTScore}. These assess the average METEOR and BERTScore values for each aligned pair $\{s_{i},s'_{i'}\}$ $\forall i$ (where ${i'}$ refers to the target index for $s_{i}$). Higher values correspond to more content preservation, where each output sentence is more likely in the correct position. 
Some drop is expected in modulating for $\pi_{i'}$, but the overall content should be faithful. 
These metrics serve more as validation, where reasonable values (e.g. $>$ 50)\footnote{Assuming the values are multiplied by 100.} are sufficient. Lower values indicate more changing of the text which may be necessary for certain narrative reorderings.

\section{Experiments}
\label{sec:experiments}

\paragraph{Model Finetuning and Generation}\label{sec:training_finetuning}\mbox{}\\
For finetuning our models, we try different combinations of learning rates (LR) for both stages. We look at either the loss (for BART and T5) or perplexity (for GPT-2) on the respective validation splits (devUnsup for 1st stage and devSup for 2nd), and choose the epoch with the lowest.

We evaluate each model on testSup, where we can directly compare results to NAREORC's human rewritings. We generate a single output per test example. The inputs are the original examples to NAR-r models 
and the $S'_{naive}$ of the examples to NAR-d models. 
See \S\ref{sec:models} for more details.

We only keep the first five sentences of each output. For BART and T5, we use beam search with a width of 5.\footnote{Nucleus sampling did not work as well for BART and T5.} For GPT-2, 
we use a nucleus sampling budget \cite{holtzman2019curious} of 0.9 and output length limit of 500. We try various softmax temperatures 
and find 0.9 performs best. 
For GPT-2, 
during finetuning, it is given the concatenation of the input plus output. During generation, it is only fed the input for which it generates a continuation (the output). We noticed that many GPT-2 generations included trailing exclamation marks, and strip these if more than four occur in a row.\footnote{See Appendix C for more finetuning/generation details.} 

\paragraph{Human Evaluation}\mbox{}\\
Annotators evaluate 100 testSup examples each from the original stories, human rewritings, outputs from our two-stage models, and a subset of one-stage models. Each example is evaluated by two annotators. 
See Appendix D for more.


They evaluate fluency, coherence, logic, and plot preservation (plot-pres) on 1-5 scales. Fluency 
is a measure of how fluent and readable a text is. Coherence is how well individual sentences fit together \cite{barzilay2008modeling}. Logic is the plausibility of described events. 
Plot-pres is how well reordered text preserves the plot of the original. This includes details about characters, events, and interactions between them, encompassing its semantic and temporal aspects.

We also conduct an interestingness (interest) study on human rewritings and outputs from our BART-2S and T5-2S models. Each reordered story's interestingness w.r.t. suspense and time flow compared to the original are evaluated from 1-5 by two annotators. 
We ask the following: \textit{``On a scale of 1-5, with 1 being most decrease in interestingness and 3 being same level of interestingness and 5 being most increase in interestingness, how interesting is the suspense and flow of time in the story S, compared to the original story O? How exciting did you find the story as you read through it?"}

\section{Results and Analysis}
\label{sec:results_and_analysis}
We present evaluation results of our 2S and subset of 1S models on testSup compared to human rewritings and original stories. 
Tables~\ref{tab:human_results_1} and \ref{tab:interestingness_results} contain human evaluation results, and Table \ref{tab:automatic_results_1} automatic evaluation results. 
Correlations between automatic and human metrics are in Table \ref{tab:correlations_table}. 
Table \ref{tab:qualitative_body_1} contains qualitative examples, with more in Appendix E.



\begin{table}[!ht]
\centering
\resizebox{\columnwidth}{!}{%
\begin{tabular}{|c|c|c|c|c|}
 \hline
 \underline{\textit{Method\textbackslash Metric}} & \textbf{Fluency} & \textbf{Coherence} & \textbf{Logic} & \textbf{Plot-pres}\\
 \hline 
  \textbf{Original stories} & 4.209 & 4.0 & 3.851 & N/A\\
 \hline
  \textbf{Human rewritings} & 3.797 & 3.723 & 3.784 & 3.972\\
 \Xhline{2\arrayrulewidth}
  \textbf{GPT2-d-2S} & 3.635 & 3.399 & 3.399 & 3.708\\
 \hline
  \textbf{GPT2-r-2S} & 3.595 & 3.378 & 3.291 & 3.375\\
 \hline
  \textbf{BART-d-1S} & 3.628 & 3.412 & 3.318 & 3.847\\
 \hline
  \textbf{BART-d-2S} & \textbf{3.818} & \underline{3.507} & 3.493 & 3.722\\
 \hline
  \textbf{BART-r-2S} & 3.757 & 3.439 & 3.493 & \underline{3.861}\\
 \hline
  \textbf{T5-d-2S} & 3.764 & 3.419 & \underline{3.5} & \textbf{3.889}\\
 \hline
  \textbf{T5-r-1S} & 3.655 & 3.378 & 3.486 & 3.847\\
 \hline
  \textbf{T5-r-2S} & \underline{3.784} & \textbf{3.595} & \textbf{3.520} & \underline{3.861}\\
 \hline
\end{tabular}}
\caption{\small Average human evaluation results on testSup (excl. interestingness), rated from 1-5. Bold corresponds to best model performance per metric, and underline second-best model performance.}
\label{tab:human_results_1}
\end{table}

\begin{table}[!ht]
\centering
\small
\begin{tabular}{|c|c|c|c|c|c|}
 \hline
 \underline{\textit{Method:}} & \textbf{Human} & \textbf{BART-d} & \textbf{BART-r} & \textbf{T5-d} & \textbf{T5-r}\\
 \hline 
  \textbf{Interest} & \textbf{3.75} & 3.367 & 3.483 & \underline{3.533} & 3.3\\
  \hline
\end{tabular}
\caption{\small Average interestingness results on testSup, rated from 1-5 (3 represents equal to original story). Models are 2S versions. 
Bold corresponds to best performance, and underline second-best.}
\label{tab:interestingness_results}
\end{table}

\begin{table*}[!ht]
\centering
\small
\begin{tabular}{|c|c|c|c|c|c|}
 \hline
 \underline{\textit{Method\textbackslash Metric}} & \textbf{BERTScore} & \textbf{BLEU} & \textbf{METEOR} & \textbf{TOF-BERTScore} & \textbf{TOF-METEOR}\\
 \hline 
  \textbf{Human rewritings} & N/A & N/A & N/A & 66.85 & 56.79\\
 \Xhline{2\arrayrulewidth}
  \textbf{GPT2-d-2S} & 60.75 & 37.01 & 45.20 & 79.23 & 74.23\\
 \hline
  \textbf{GPT2-r-2S} & 58.03 & 32.57 & 40.85 & 73.04 & 63.00\\
 \hline
  \textbf{BART-d-1S} & 67.14 & 44.73 & 49.88 & 95.61 & 93.43\\
 \hline
  \textbf{BART-d-2S} & \underline{67.93} & \underline{46.03} & \underline{50.54} & 93.55 & 90.81\\
 \hline
  \textbf{BART-r-2S} & 67.16 & 44.63 & 49.16 & 91.32 & 86.43\\
 \hline
  \textbf{T5-d-2S} & \textbf{67.99} & \textbf{46.95} & \textbf{51.12} & 94.20 & 91.83\\
 \hline
  \textbf{T5-r-1S} & 66.24 & 43.40 & 48.20 & 89.85 & 84.26\\
 \hline
  \textbf{T5-r-2S} & 66.62 & 44.30 & 49.00 & 91.61 & 86.16\\
 \hline
\end{tabular}
\caption{\small Average automatic evaluation results on testSup (values multiplied by 100). Bold corresponds to best performance per metric, and underline second-best (excluding the TOF metrics which are mainly for validation).}
\label{tab:automatic_results_1}
\end{table*}

\begin{table*}[!ht]
\centering
\small
\begin{tabular}{|c|c|c|c|c|c|c|}
 \cline{1-7}
 \underline{\textit{Metric}} & \underline{\textit{Correlation}} & \textbf{Fluency} & \textbf{Coherence} & \textbf{Logic} & \textbf{Plot-pres} & \textbf{Interest}\\
 \hline
  \multirow{2}{*}{\textbf{BERTScore}} & Pearson & 0.130 (4e-04) & 0.139 (1e-04) & \textbf{0.125} (0.001) & \textbf{0.255} (1e-06) & 0.111 (0.226)\\ %
 \cline{2-7}
  & Spearman & 0.106 (0.004) & 0.124 (0.001) & \textbf{0.127} (0.001) & \textbf{0.211} (5e-05) & 0.117 (0.201)\\
 \Xhline{2\arrayrulewidth}
  \multirow{2}{*}{\textbf{BLEU}} & Pearson & \textbf{0.144} (9e-05) & \textbf{0.140} (1e-04) & 0.113 (0.002) & 0.219 (3e-05) & \textbf{0.174} (0.047)\\
 \cline{2-7}
  & Spearman & \textbf{0.130} (4e-04) & \textbf{0.129} (4e-04) & 0.123 (0.001) & 0.179 (0.001) & \textbf{0.171} (0.049)\\
 \Xhline{2\arrayrulewidth}
  \multirow{2}{*}{\textbf{METEOR}} & Pearson & 0.107 (0.003) & 0.125 (0.001) & 0.108 (0.003) & 0.203 (1e-04) & 0.120 (0.191)\\
 \cline{2-7}
  & Spearman & 0.098 (0.008) & 0.114 (0.002) & 0.122 (0.001) & 0.164 (0.002) & 0.121 (0.187)\\
 \hline
\end{tabular}
\caption{\small Pearson and Spearman correlations between automatic and human evaluation metrics, with p-values in brackets. TOF metrics excluded as they are mainly for validation. Bold corresponds to highest correlation per human evaluation metric.}
\label{tab:correlations_table}
\end{table*}

\begin{table*}[ht!]
\centering
\small
\addtolength{\tabcolsep}{-4pt}
\begin{tabular}{|p{1.1cm}|p{16.3cm}|}
\hline 
\textbf{Method} & \textbf{Text} \\ \hline 
Original \newline (ex. 1) & {Jimmy wandered around town looking for a soda. Before he knew it, he was in a strange area. He was scared of strangers and didn't want to ask anyone. Soon a policeman came by and asked if he was lost. He told him that he was lost.} \{5,4,2,1,3\} \\ \hline
BART-d & {{\textbf{Jimmy told a policeman} that he was lost. The policeman \textbf{had come} by and asked if he \textbf{had been} lost. Before he knew it, he was in a strange area. Jimmy had wandered around town looking for a soda. He was scared of strangers and didn't want to ask anyone.}} \\ \hline 
BART-r & {{\textbf{Jimmy told the policeman} that he was lost. The policeman came by and asked if he had lost \textbf{his wallet}. Before Jimmy knew it, he was in a strange area. He had wandered around town looking for a soda. He was scared of strangers and didn't want to ask anyone.}} \\ \hline 
T5-d & {{\textbf{Jimmy told a policeman} that he was lost. \textbf{Soon after}, a policeman came by and asked if he was lost. Before he knew it, he was in a strange area. He had wandered around town looking for a soda. He was scared of strangers and didn't want to ask anyone.}} \\ \hline 
T5-r & {{\textbf{Jimmy told a policeman} that he was lost. A policeman came by and asked \textbf{if he had a soda}. Before he knew it, he was in a strange area. He had wandered around town looking for a soda. He was scared of strangers and didn't want to ask anyone.}} \\ \Xhline{3\arrayrulewidth}
Original \newline (ex. 2) & {Every morning, a bird woke Fred up for work at 6 AM. Fred began to like the bird. And he enjoyed the sound of its singing. One day, the bird no longer sang. He hoped nothing bad had happened to the bird.} \{4,1,3,5,2\} \\ \hline 
Human & {{Fred didn't hear the bird singing this morning which is unusual. The bird had been waking him up every single day at 6 AM for quite some time. He had \textbf{grown accustomed to the bird being his alarm clock}. Now he's worried that something might have happened to the poor bird. He almost considers the bird a friend at this point.}} \\ \hline 
GPT2-r & {{One day, Fred's bird began to sing. Every morning, \textbf{Fred woke him up} for work at 6 AM. \textbf{This was because he liked Fred}. He hoped nothing bad had happened to Fred. \textbf{This bird, however, did not like Fred}.}} \\ \hline 
BART-r & {{\textbf{Fred} no longer sang. Every morning, he woke Fred up for work at 6 AM. He enjoyed the sound of his singing. He hoped nothing bad had happened to him. He \textbf{had begun} to like the bird.}} \\ \hline 
T5-d & {{One day, the bird no longer sang. Every morning, a bird woke Fred up for work at 6 AM. He \textbf{had enjoyed} the sound of its singing. He had hoped nothing bad had happened to the bird. \textbf{After a while}, Fred began to like the bird.}} \\ \Xhline{3\arrayrulewidth}
Original \newline (ex. 3) & {Eric was walking around a pond at a local park. He saw a turtle on a rock few feet offshore. He stepped into the water to try to catch the turtle. Suddenly, he sank knee deep into the mud and got stuck. Someone walking on the shore ran over to rescue Eric.} \{5,4,1,3,2\} \\ \hline 
Human & {{Someone who had been walking on the shore ultimately ran over and rescued Eric from the mud. Eric's knee had sunk deep into the mud, rendering him stuck. Earlier, Eric had been walking around a pond while at a local park. Trying to catch a turtle in the pond, Eric stepped into the water. Eric did so because he had seen a turtle poached on a rock few feet offshore.}} \\ \hline 
BART-d & {{Someone walking on the shore ran over to rescue Eric. \textbf{Earlier}, he had sunk knee deep into the mud and got stuck. Eric was walking around a pond at a local park. He stepped into the water to try to catch the turtle. He \textbf{had seen} a turtle on a rock few feet offshore.}} \\ \hline
BART-r & {{Someone walking on the shore ran over to rescue Eric. Eric sank knee deep into the mud and got stuck. He was walking around a pond at a local park. He \textbf{had stepped} into the water to try to catch the turtle. He saw the turtle few feet offshore.}} \\ \hline 
T5-r & {{Someone walking on the shore ran over to rescue Eric. \textbf{Suddenly}, Eric sank knee deep into the mud and got stuck. He was walking around a pond at a local park. He stepped into the water to try to catch the turtle. He saw the turtle few feet offshore.}} \\ \Xhline{3\arrayrulewidth}
Original \newline (ex. 4) & {Joey went to Disneyland with his family. He was so excited to meet Mickey Mouse. When he met Mickey Mouse he was so excited he couldn't speak! Joey finally said hi to Mickey and it was great! Joey had a great time.} \{5,2,4,1,3\} \\ \hline 
GPT2-d & {{Joey had a great time. He was so excited to meet Mickey Mouse. He had finally said hi to Mickey and it was great! He had gone to Disneyland with his family. When he met Mickey Mouse he was so excited he couldn't speak}} \\ \hline 
GPT2-r & {{Joey had a great time on the \textbf{roller coaster} at Disneyland. \textbf{He was so excited} to meet Mickey Mouse he couldn't speak! 
(...) \textbf{He was so excited} that he went to the park with his family. \textbf{He was so excited} when he met Mickey Mouse and \textbf{he was so excited}.}} \\ \hline
BART-r & {{Joey had a great time \textbf{at Disneyland}. He was so excited to meet Mickey Mouse. He finally said hi to Mickey and it was great! He had gone to Disneyland with his family. When he met Mickey he was excited he couldn't speak!}} \\ \hline
T5-d & {{Joey had a great time \textbf{at Disneyland}. He was so excited to meet Mickey Mouse. He had finally said hi to Mickey and it was great! He had gone to Disneyland with his family. When he met Mickey Mouse he was so excited he couldn't speak!}} \\ \hline
\end{tabular}
\caption{\small Qualitative examples for testSup. Target perms are in brackets beside original stories. \textit{d} refers to denoise, and \textit{r} to reorder (2S models).}
\label{tab:qualitative_body_1}
\end{table*}

\subsection{Analysis of Human Evaluation Results}\label{sec:human_eval_analysis}
We begin by analyzing human evaluation performance through results in Tables~\ref{tab:human_results_1} and \ref{tab:interestingness_results}.

\paragraph{Fluency, Coherence, Logic:} Original stories are the highest for all three metrics\footnote{Although these metrics slightly decrease for reordered stories, we note that NAREOR's main purpose is for more interesting tellings of the same story which we do achieve (see Table \ref{tab:interestingness_results}).} with human rewritings second for coherence and logic, beating the models by a noticeable degree. BART-d-2S and T5-r-2S are generally the best-performing models here. BART-d-2S slightly outperforms human rewritings on fluency, with T5-r-2S closely behind, demonstrating that these models are quite fluent. These models also outdo their 1S variants. 
GPT-2 models perform worst on all metrics. 

\paragraph{Plot-pres:} We see that human rewritings best preserve the plot of the original stories. T5-d-2S is the best performing model on plot-pres, followed by BART-r-2S and T5-r-2S. 
GPT-2 models perform the worst 
at preserving the plot of the original stories (which we show qualitatively in \S\ref{subsec:qualitative_analysis}).

\paragraph{Interestingness:}\label{sec:interest_results} Human rewritings 
score highest on interest. 
Humans rewrite the text in more creative ways, whereas BART and T5 models are more conservative (see \S\ref{sec:TOF_analysis} TOF and \S\ref{subsec:qualitative_analysis}). Narrative reorderings for all methods are more interesting, on average, than original stories. NAREOR can indeed be used to generate more interesting story variations.


\subsection{Analysis of Automatic Evaluation Results}\label{sec:automatic_eval_analysis}
We now analyze the automatic evaluation performance of the different methods in Table~\ref{tab:automatic_results_1}.

\paragraph{BERTScore, BLEU, METEOR:} We see from Table \ref{tab:correlations_table} that these reference-based metrics correlate quite well with human eval metrics, particularly plot-pres. 
T5-d-2S performs best followed by BART-d-2S. Similar to the human evaluation, 2S models outperform their 1S variants, and GPT-2 models perform worst overall. 
Denoise outperforms reorder variants and generate more similar text, on avg, to human references.

\paragraph{Target Order Fidelity (TOF):}\label{sec:TOF_analysis} It appears all approaches are reasonable (e.g. $>$ 50 for TOF metrics), and outputs are likely in the correct target orders. 
Human rewritings have the lowest TOF; humans are less conservative while rewriting 
(shown in \S\ref{subsec:qualitative_analysis}). GPT-2 models modify text second heaviest, but perform worst overall. 
They introduce more errors, e.g. repeating or hallucinating to degrade text quality and plot-pres (\S\ref{subsec:qualitative_analysis}). BART and T5 models are more conservative. It appears they have learned to perform minimal but effective edits (\S\ref{subsec:qualitative_analysis}). They lag behind humans 
and heavier editing may be required to further improve. Lastly, it appears the reorder models modify text more heavily than their denoise variants. 

\subsection{Qualitative Analysis} \label{subsec:qualitative_analysis}
From Table \ref{tab:qualitative_body_1}, we see that humans modify text heavily to suit the reorderings and are sometimes quite creative, e.g. phrasing Fred as having \textit{grown accustomed to the bird being his alarm clock} (ex. 2). Humans successfully handle necessary coreferences, tenses, time expressions (timexes), etc. 

GPT-2 modifies text quite heavily but suffers from incorrect coreference while introducing spurious tokens, repetition, or hallucations. 
For ex. 2, GPT2-r changes the plot greatly, stating \textit{Fred woke him up for work} and \textit{This was because he liked Fred} (likely due to poor coreference), and hallucinating \textit{This bird, however, did not like Fred}. For ex. 4, it repeats Joey's excitement many times, while hallucinating a \textit{roller coaster} that was absent in the original story.

BART and T5 models are more conservative, but their edits are important and effective. They handle coreference, tense, and timexes quite well. These pinpointed and critical edits are required to maintain plot. For ex. 1, they modify \textit{He told him that he was lost} to \textit{\textbf{Jimmy} told \textbf{a/the policeman} that he was lost} given that sentence is now at the beginning. BART-d impressively modifies tense by converting \textit{Soon a policeman came by and asked if he was lost} to \textit{The policeman \textbf{had come} by and asked if he \textbf{had been} lost}. For ex. 2, T5-d converts \textit{enjoyed} to \textit{had enjoyed} since the bird no longer singing is now prior information, and adds the timex \textit{After a while} to the beginning of the last output sentence. BART-r successfully changes \textit{Fred began to like the bird} to \textit{He \textbf{had begun} to like the bird}. For ex. 3, BART-d inserts the timex \textit{Earlier} at the beginning of the second output sentence, correctly and unambiguously conveying its underlying temporality w.r.t. the first. BART-d correctly changes \textit{saw a turtle} to \textit{\textbf{had seen} a turtle}, while BART-r does so for \textit{stepped} to \textit{had stepped}. For ex. 4, BART and T5 models all resolve the \textit{Disneyland} ellipsis by converting \textit{Joey had a great time} to \textit{Joey had a great time \textbf{at Disneyland}}, while GPT2-d cannot.

However, the BART and T5 models are imperfect. For ex. 1, BART-r hallucatines \textit{lost his wallet} (original story does not involve a wallet), T5-d inserts an incorrect timex of \textit{Soon after} at the beginning of the second output sentence, and T5-r hallucinates \textit{asked if he had a soda} (
this is not asked in the original story). For ex. 2, BART-r incorrectly converts \textit{\textbf{the bird} no langer sang} to \textit{\textbf{Fred} no longer sang}, likely due to coreference difficulties. For ex. 3, T5-r does not convert \textit{Suddenly} to \textit{Earlier} like BART-d, giving a false interpretation that Eric slipped after his rescuer's arrival. BART-r does not mislead with \textit{Suddenly}, but is ambiguous with no timex at all. Further, BART and T5 are more conservative than humans.

\subsection{Overall Takeaways}
Humans modify text greatly while successfully performing NAREOR. BART and T5 models perform decently with minimal but effective edits. GPT-2 models tend to repeat, hallucinate, and reduce text quality and plot preservation.

 
Based on human (\S\ref{sec:human_eval_analysis}) and automatic (\S\ref{sec:automatic_eval_analysis}) evaluation, 
BART-d-2S and T5-d-2S are the best models overall. BART-d-2S outdoes its reorder variant, possibly due to BART's pretraining as a denoising autoencoder, closer to our denoise training method. For T5, both methods perform quite well and show potential. However, T5-d outperforms on plot-pres (Table \ref{tab:human_results_1}), interest (Table \ref{tab:interestingness_results}), and automatic metrics (Table \ref{tab:automatic_results_1}). The denoise training method appears to be slightly more effective, possibly because it is partially inspired by how humans perform NAREOR (see \S\ref{sec:models}). These are the first two task-specific training methods for NAREOR which we propose ourselves, each approaching the task differently (see \S\ref{sec:models}). 2S models also mostly outperform 1S ones, demonstrating that second stage finetuning improves upon the first. 

BART and T5 models are quite effective, excelling at fluency, but have further room for improvement in coherence, logic, plot-pres, and interest. \S\ref{subsec:qualitative_analysis} shows they still suffer from several issues. Their conservative tendency may limit their NAREOR ability compared to humans. 
Overall, these models serve as strong initial baselines for NAREOR while underscoring the task's difficulty and potential for exploration.

\section{Applications of NAREOR}
\label{sec:applications}
Sentence ordering involves reconstructing original sentence order of an unordered sentence set \cite{barzilay2008modeling}. 
NAREORC's reordered stories could serve as a challenge set for 
sentence reordering models due to their non-linear narrative structure underrepresented in  
corpora. We use the implementation of \citet{prabhumoye2020topological} to train i) $M_{ext}$, an external model on the SIS corpus \cite{ferraro2016visual}, ii) $M_{iid}$, an in-domain model on first 20\% of ROCStories' train split. We test each on i) Control set $\{s_{i}\}_{i=1}^{i=n}$, input stories from testSup, ii) Challenge set $\{s'_{i}\}_{i=1}^{i=n}$, reordered stories from testSup. Table \ref{tab:applications_table} shows drastic drops across metrics (higher is better - see \citet{prabhumoye2020topological}) for both $M_{ext}$ and $M_{iid}$ from control to challenge set, confirming our hypothesis.

\begin{table}[!ht]
\centering
\resizebox{\columnwidth}{!}{
\begin{tabular}{|c|c|c|c|c|c|}
 \cline{1-6}
 \underline{\textit{Model}} & \underline{\textit{TestSet}} & \textbf{SentAcc} & \textbf{Rouge-S} & \textbf{LCS} & \textbf{Kendall $\tau$}\\
 \hline
  \multirow{2}{*}{\textbf{$M_{ext}$}} & Control & 76.35 & 48 & 59.1 & 0.57 \\ %
 \cline{2-6}
  & Challenge & 52.4 & 24.7 & 29.7 & 0.12\\
 \Xhline{2\arrayrulewidth}
  \multirow{2}{*}{\textbf{$M_{iid}$}} & Control & 66.4 & 85.3 & 84.8 & 0.75\\
 \cline{2-6}
  & Challenge & 21.9 & 49.6 & 58 & 0.03\\
 \hline
\end{tabular}}
\caption{ \small Sentence ordering on control vs. challenge sets.}
\label{tab:applications_table}
\end{table}

Systems with ability to manipulate 
narrative variables like order could be important for automating pedagogical setups, especially for fine-grained language skills such as \emph{argumentation in essay writing}. As \citet{wingate2012argument} explains, tutor understanding is found deficient and methods of feedback for students are inconsistent or vague. Language in school texts follows a characteristic register, which often differs from registers students handle in everyday conversation \cite{schleppegrell2001linguistic}. Models (e.g. NAREOR ones) which can control elements of register, e.g narrative order, can be used to tailor such content to intended settings and bridge this gap.

Systems that can generate event timelines for clinical narratives, e.g. admission notes and physical reports, is important for applications like 
medical document summarization \cite{bramsen-etal-2006-inducing,Reichert2010CognitiveAO} and clinical decision making \cite{DEMNERFUSHMAN2009760}. \citet{raghavan-etal-2014-cross} demonstrate that cross-narrative temporal ordering of medical events is vital to generating a comprehensive timeline over a patient's history. Aligning multiple medical event sequences using 
coreference information and temporal relations has a large impact on their presentation and effectiveness. Our NAREOR models may be effective here and improve upon existing systems.

\section{Related Work}
\label{sec:related}
There exists work on the sentence ordering task discussed in \S\ref{sec:applications}. 
For example, \citet{chen2016neural} learn pairwise orderings of sentences using a ranking model. 
Unlike sentence ordering, NAREOR involves reordering and rewriting a sequence of sentences to fit a new narrative order. 

TALESPIN \cite{meehan-1975-using} was an early goal-based story generator. There has since been work on related tasks like story cloze test \cite{mostafazadeh-etal-2016-generating,mostafazadeh-etal-2017-lsdsem} and generation from prompts \cite{fan2018hierarchical,see2019massively}. 
Some works explore controllable variants, e.g. with keywords as control \cite{peng2018towards}. NAREOR is distinct as it aims to preserve the underlying plot while controlling a story-level aspect for an already-complete story.



\citet{piper-etal-2021-narrative} 
situate NLP research within narratological traditions, linking computational NLP work to theory. \S2.4 of their paper examines the temporality \textit{element of narrativity}, which we investigate the manipulation of. \citet{kim2017visualizing} annotate story order for movie scripts and visualize narrative order as a function of story order.

There have also been several works in the broader controllable and constrained text generation space. \citet{miao2019cgmh} use Metropolis-Hastings sampling to determine Levenshtein edits per generation step. \citet{feng2019keep} propose Semantic Text Exchange to adjust topic-level text semantics. Generative commonsense reasoning, or CommonGen \cite{lin-etal-2020-commongen}, is a concept-to-text generation task where the goal is to generate a fluent and logical sentence given a set of input concepts or keywords. Many works investigate ways to improve seq2seq model performance on this task, including EKI-BART \cite{fan2020enhanced}, KG-BART \cite{liu2020kg}, SAPPHIRE \cite{feng-etal-2021-sapphire}, VisCTG \cite{VisCTG}, and RE-T5 \cite{wang2021retrieval}.


\section{Conclusion and Future Work}
\label{sec:conclusions}
We proposed the task of Narrative Reordering (NAREOR) and introduced a dataset, NAREORC, 
with task-specific training methods and evaluation metrics, and experimented with T5, BART, and GPT-2. Extensive evaluation and qualitative analysis demonstrated that our models are quite effective but can be further improved, and that NAREOR is challenging with potential for further exploration. We showed that NAREOR can be used to produce more interesting story variations  and as a challenge set for tasks like sentence ordering.


Future directions include exploring training ideas better emulating human rewrites. 
NAREOR can be investigated as document-level paraphrasing for applications like data augmentation \cite{feng2021survey}
for document tasks and text generation \cite{feng-etal-2020-genaug}, 
and applications for education and medicine (\S\ref{sec:applications}). More challenging task variations (e.g. sub-sentential) can also be explored. Lastly, NAREOR's controllability aspect can be further investigated.
\bibliography{anthology,aaai22}

\begin{thebibliography}{52}
\providecommand{\natexlab}[1]{#1}

\bibitem[{Banerjee and Lavie(2005)}]{banerjee2005meteor}
Banerjee, S.; and Lavie, A. 2005.
\newblock {METEOR}: An automatic metric for {MT} evaluation with improved
  correlation with human judgments.
\newblock In \emph{Proceedings of the acl workshop on intrinsic and extrinsic
  evaluation measures for machine translation and/or summarization}, 65--72.

\bibitem[{Barzilay and Lapata(2008)}]{barzilay2008modeling}
Barzilay, R.; and Lapata, M. 2008.
\newblock Modeling local coherence: An entity-based approach.
\newblock \emph{Computational Linguistics}, 34(1): 1--34.

\bibitem[{Bos(1995)}]{bos1995jewish}
Bos, G. 1995.
\newblock Jewish {T}raditions on {S}trengthening {M}emory and {L}eone
  {M}odena's Evaluation.
\newblock \emph{Jewish Studies Quarterly}, 2(1): 39--58.

\bibitem[{Bramsen et~al.(2006)Bramsen, Deshpande, Lee, and
  Barzilay}]{bramsen-etal-2006-inducing}
Bramsen, P.; Deshpande, P.; Lee, Y.~K.; and Barzilay, R. 2006.
\newblock Inducing Temporal Graphs.
\newblock In \emph{Proceedings of the 2006 Conference on Empirical Methods in
  Natural Language Processing}, 189--198. Sydney, Australia: Association for
  Computational Linguistics.

\bibitem[{Chang and Manning(2012)}]{chang2012sutime}
Chang, A.~X.; and Manning, C.~D. 2012.
\newblock Sutime: {A} library for recognizing and normalizing time expressions.
\newblock In \emph{LREC}, volume 2012, 3735--3740.

\bibitem[{Chen, Qiu, and Huang(2016)}]{chen2016neural}
Chen, X.; Qiu, X.; and Huang, X. 2016.
\newblock Neural sentence ordering.
\newblock \emph{arXiv preprint arXiv:1607.06952}.

\bibitem[{Demner-Fushman, Chapman, and McDonald(2009)}]{DEMNERFUSHMAN2009760}
Demner-Fushman, D.; Chapman, W.~W.; and McDonald, C.~J. 2009.
\newblock What can natural language processing do for clinical decision
  support?
\newblock \emph{Journal of Biomedical Informatics}, 42(5): 760--772.
\newblock Biomedical Natural Language Processing.

\bibitem[{Elder et~al.(2018)Elder, Gehrmann, O’Connor, and
  Liu}]{elder2018e2e}
Elder, H.; Gehrmann, S.; O’Connor, A.; and Liu, Q. 2018.
\newblock E2E nlg challenge submission: {T}owards controllable generation of
  diverse natural language.
\newblock In \emph{Proceedings of the 11th International Conference on Natural
  Language Generation}, 457--462.

\bibitem[{Fan, Lewis, and Dauphin(2018)}]{fan2018hierarchical}
Fan, A.; Lewis, M.; and Dauphin, Y. 2018.
\newblock Hierarchical Neural Story Generation.
\newblock In \emph{Proceedings of the 56th Annual Meeting of the Association
  for Computational Linguistics (Volume 1: Long Papers)}, 889--898.

\bibitem[{Fan et~al.(2020)Fan, Gong, Wei, Wang, Huang, Jiao, Huang, Duan, and
  Zhang}]{fan2020enhanced}
Fan, Z.; Gong, Y.; Wei, Z.; Wang, S.; Huang, Y.; Jiao, J.; Huang, X.; Duan, N.;
  and Zhang, R. 2020.
\newblock An Enhanced Knowledge Injection Model for Commonsense Generation.
\newblock In \emph{Proceedings of the 28th International Conference on
  Computational Linguistics}, 2014--2025. Barcelona, Spain (Online):
  International Committee on Computational Linguistics.

\bibitem[{Feng et~al.(2020)Feng, Gangal, Kang, Mitamura, and
  Hovy}]{feng-etal-2020-genaug}
Feng, S.~Y.; Gangal, V.; Kang, D.; Mitamura, T.; and Hovy, E. 2020.
\newblock {G}en{A}ug: Data Augmentation for Finetuning Text Generators.
\newblock In \emph{Proceedings of Deep Learning Inside Out (DeeLIO): The First
  Workshop on Knowledge Extraction and Integration for Deep Learning
  Architectures}, 29--42. Online: Association for Computational Linguistics.

\bibitem[{Feng et~al.(2021{\natexlab{a}})Feng, Gangal, Wei, Chandar, Vosoughi,
  Mitamura, and Hovy}]{feng2021survey}
Feng, S.~Y.; Gangal, V.; Wei, J.; Chandar, S.; Vosoughi, S.; Mitamura, T.; and
  Hovy, E. 2021{\natexlab{a}}.
\newblock A Survey of Data Augmentation Approaches for NLP.
\newblock \emph{ACL 2021 Findings}.

\bibitem[{Feng et~al.(2021{\natexlab{b}})Feng, Huynh, Narisetty, Hovy, and
  Gangal}]{feng-etal-2021-sapphire}
Feng, S.~Y.; Huynh, J.; Narisetty, C.~P.; Hovy, E.; and Gangal, V.
  2021{\natexlab{b}}.
\newblock {SAPPHIRE}: Approaches for Enhanced Concept-to-Text Generation.
\newblock In \emph{Proceedings of the 14th International Conference on Natural
  Language Generation}, 212--225. Aberdeen, Scotland, UK: Association for
  Computational Linguistics.

\bibitem[{Feng, Li, and Hoey(2019)}]{feng2019keep}
Feng, S.~Y.; Li, A.~W.; and Hoey, J. 2019.
\newblock {K}eep {C}alm and {S}witch {O}n! {P}reserving {S}entiment and
  {F}luency in {S}emantic {T}ext {E}xchange.
\newblock In \emph{Proceedings of the 2019 Conference on Empirical Methods in
  Natural Language Processing and the 9th International Joint Conference on
  Natural Language Processing (EMNLP-IJCNLP)}, 2701--2711.

\bibitem[{Feng et~al.(2022)Feng, Lu, Tao, Alikhani, Mitamura, Hovy, and
  Gangal}]{VisCTG}
Feng, S.~Y.; Lu, K.; Tao, Z.; Alikhani, M.; Mitamura, T.; Hovy, E.; and Gangal,
  V. 2022.
\newblock Retrieve, Caption, Generate: Visual Grounding for Enhancing
  Commonsense in Text Generation Models.
\newblock \emph{AAAI 2022}.

\bibitem[{Genette(1983)}]{genette1983narrative}
Genette, G. 1983.
\newblock \emph{Narrative discourse: An essay in method}, volume~3.
\newblock Cornell University Press.

\bibitem[{Halliwell et~al.(1998)}]{halliwell1998aristotle}
Halliwell, S.; et~al. 1998.
\newblock \emph{{A}ristotle's poetics}.
\newblock University of Chicago Press.

\bibitem[{Holtzman et~al.(2019)Holtzman, Buys, Du, Forbes, and
  Choi}]{holtzman2019curious}
Holtzman, A.; Buys, J.; Du, L.; Forbes, M.; and Choi, Y. 2019.
\newblock The Curious Case of Neural Text Degeneration.
\newblock In \emph{International Conference on Learning Representations}.

\bibitem[{Huang et~al.(2016)Huang, Ferraro, Mostafazadeh, Misra, Agrawal,
  Devlin, Girshick, He, Kohli, Batra, Zitnick, Parikh, Vanderwende, Galley, and
  Mitchell}]{ferraro2016visual}
Huang, T.-H.~K.; Ferraro, F.; Mostafazadeh, N.; Misra, I.; Agrawal, A.; Devlin,
  J.; Girshick, R.; He, X.; Kohli, P.; Batra, D.; Zitnick, C.~L.; Parikh, D.;
  Vanderwende, L.; Galley, M.; and Mitchell, M. 2016.
\newblock Visual Storytelling.
\newblock In \emph{Proceedings of the 2016 Conference of the North {A}merican
  Chapter of the Association for Computational Linguistics: Human Language
  Technologies}, 1233--1239. San Diego, California: Association for
  Computational Linguistics.

\bibitem[{HuggingFace(2020)}]{neuralcoref}
HuggingFace. 2020.
\newblock NeuralCoref.
\newblock [Online; accessed 29-September-2020].

\bibitem[{Kendall(1938)}]{kendall1938new}
Kendall, M.~G. 1938.
\newblock A new measure of rank correlation.
\newblock \emph{Biometrika}, 30(1/2): 81--93.

\bibitem[{Keskar et~al.(2019)Keskar, McCann, Varshney, Xiong, and
  Socher}]{keskar2019ctrl}
Keskar, N.~S.; McCann, B.; Varshney, L.~R.; Xiong, C.; and Socher, R. 2019.
\newblock Ctrl: A conditional transformer language model for controllable
  generation.
\newblock \emph{arXiv preprint arXiv:1909.05858}.

\bibitem[{Kim et~al.(2017)Kim, Bach, Im, Schriber, Gross, and
  Pfister}]{kim2017visualizing}
Kim, N.~W.; Bach, B.; Im, H.; Schriber, S.; Gross, M.; and Pfister, H. 2017.
\newblock Visualizing nonlinear narratives with story curves.
\newblock \emph{IEEE transactions on visualization and computer graphics},
  24(1): 595--604.

\bibitem[{Lewis et~al.(2020)Lewis, Liu, Goyal, Ghazvininejad, Mohamed, Levy,
  Stoyanov, and Zettlemoyer}]{lewis-etal-2020-bart}
Lewis, M.; Liu, Y.; Goyal, N.; Ghazvininejad, M.; Mohamed, A.; Levy, O.;
  Stoyanov, V.; and Zettlemoyer, L. 2020.
\newblock {BART}: Denoising Sequence-to-Sequence Pre-training for Natural
  Language Generation, Translation, and Comprehension.
\newblock In \emph{Proceedings of the 58th Annual Meeting of the Association
  for Computational Linguistics}, 7871--7880. Online: Association for
  Computational Linguistics.

\bibitem[{Lin et~al.(2020)Lin, Zhou, Shen, Zhou, Bhagavatula, Choi, and
  Ren}]{lin-etal-2020-commongen}
Lin, B.~Y.; Zhou, W.; Shen, M.; Zhou, P.; Bhagavatula, C.; Choi, Y.; and Ren,
  X. 2020.
\newblock {C}ommon{G}en: A Constrained Text Generation Challenge for Generative
  Commonsense Reasoning.
\newblock In \emph{Findings of the Association for Computational Linguistics:
  EMNLP 2020}, 1823--1840. Online: Association for Computational Linguistics.

\bibitem[{Liu et~al.(2021)Liu, Wan, He, Peng, and Yu}]{liu2020kg}
Liu, Y.; Wan, Y.; He, L.; Peng, H.; and Yu, P.~S. 2021.
\newblock KG-BART: Knowledge Graph-Augmented BART for Generative Commonsense
  Reasoning.
\newblock \emph{Proceedings of the AAAI Conference on Artificial Intelligence},
  35(7): 6418--6425.

\bibitem[{Meehan(1975)}]{meehan-1975-using}
Meehan, J.~R. 1975.
\newblock Using Planning Structures to Generate Stories.
\newblock \emph{American Journal of Computational Linguistics}, 78--94.
\newblock Microfiche 33.

\bibitem[{Miao et~al.(2019)Miao, Zhou, Mou, Yan, and Li}]{miao2019cgmh}
Miao, N.; Zhou, H.; Mou, L.; Yan, R.; and Li, L. 2019.
\newblock Cgmh: Constrained sentence generation by metropolis-hastings
  sampling.
\newblock In \emph{Proceedings of the AAAI Conference on Artificial
  Intelligence}, volume~33, 6834--6842.

\bibitem[{Montfort(2007)}]{montfort2007ordering}
Montfort, N. 2007.
\newblock Ordering Events in Interactive Fiction Narratives.
\newblock In \emph{AAAI Fall Symposium: Intelligent Narrative Technologies},
  87--94.

\bibitem[{Morgan(2017)}]{morgan2017narrative}
Morgan, M.~S. 2017.
\newblock Narrative ordering and explanation.
\newblock \emph{Studies in History and Philosophy of Science Part A}, 62:
  86--97.

\bibitem[{Mostafazadeh et~al.(2016{\natexlab{a}})Mostafazadeh, Chambers, He,
  Parikh, Batra, Vanderwende, Kohli, and Allen}]{mostafazadeh2016corpus}
Mostafazadeh, N.; Chambers, N.; He, X.; Parikh, D.; Batra, D.; Vanderwende, L.;
  Kohli, P.; and Allen, J. 2016{\natexlab{a}}.
\newblock A corpus and cloze evaluation for deeper understanding of commonsense
  stories.
\newblock In \emph{Proceedings of the 2016 Conference of the North American
  Chapter of the Association for Computational Linguistics: Human Language
  Technologies}, 839--849.

\bibitem[{Mostafazadeh et~al.(2016{\natexlab{b}})Mostafazadeh, Misra, Devlin,
  Mitchell, He, and Vanderwende}]{mostafazadeh-etal-2016-generating}
Mostafazadeh, N.; Misra, I.; Devlin, J.; Mitchell, M.; He, X.; and Vanderwende,
  L. 2016{\natexlab{b}}.
\newblock Generating Natural Questions About an Image.
\newblock In \emph{Proceedings of the 54th Annual Meeting of the Association
  for Computational Linguistics (Volume 1: Long Papers)}, 1802--1813. Berlin,
  Germany: Association for Computational Linguistics.

\bibitem[{Mostafazadeh et~al.(2017)Mostafazadeh, Roth, Louis, Chambers, and
  Allen}]{mostafazadeh-etal-2017-lsdsem}
Mostafazadeh, N.; Roth, M.; Louis, A.; Chambers, N.; and Allen, J. 2017.
\newblock {LSDS}em 2017 Shared Task: The Story Cloze Test.
\newblock In \emph{Proceedings of the 2nd Workshop on Linking Models of
  Lexical, Sentential and Discourse-level Semantics}, 46--51. Valencia, Spain:
  Association for Computational Linguistics.

\bibitem[{Papineni et~al.(2002)Papineni, Roukos, Ward, and
  Zhu}]{papineni-etal-2002-bleu}
Papineni, K.; Roukos, S.; Ward, T.; and Zhu, W.-J. 2002.
\newblock {B}leu: a Method for Automatic Evaluation of Machine Translation.
\newblock In \emph{Proceedings of the 40th Annual Meeting of the Association
  for Computational Linguistics}, 311--318. Philadelphia, Pennsylvania, USA:
  Association for Computational Linguistics.

\bibitem[{Peng et~al.(2018)Peng, Ghazvininejad, May, and
  Knight}]{peng2018towards}
Peng, N.; Ghazvininejad, M.; May, J.; and Knight, K. 2018.
\newblock Towards controllable story generation.
\newblock In \emph{Proceedings of the First Workshop on Storytelling}, 43--49.

\bibitem[{Piper, So, and Bamman(2021)}]{piper-etal-2021-narrative}
Piper, A.; So, R.~J.; and Bamman, D. 2021.
\newblock Narrative Theory for Computational Narrative Understanding.
\newblock In \emph{Proceedings of the 2021 Conference on Empirical Methods in
  Natural Language Processing}, 298--311. Online and Punta Cana, Dominican
  Republic: Association for Computational Linguistics.

\bibitem[{Prabhumoye, Salakhutdinov, and
  Black(2020)}]{prabhumoye2020topological}
Prabhumoye, S.; Salakhutdinov, R.; and Black, A.~W. 2020.
\newblock Topological Sort for Sentence Ordering.
\newblock In \emph{Proceedings of the 58th Annual Meeting of the Association
  for Computational Linguistics}, 2783--2792.

\bibitem[{Propp(2010)}]{propp2010morphology}
Propp, V. 2010.
\newblock \emph{Morphology of the Folktale}, volume~9.
\newblock University of Texas Press.

\bibitem[{Radford et~al.(2019)Radford, Wu, Child, Luan, Amodei, and
  Sutskever}]{radford2019language}
Radford, A.; Wu, J.; Child, R.; Luan, D.; Amodei, D.; and Sutskever, I. 2019.
\newblock Language models are unsupervised multitask learners.
\newblock \emph{OpenAI Blog}, 1(8): 9.

\bibitem[{Raffel et~al.(2020)Raffel, Shazeer, Roberts, Lee, Narang, Matena,
  Zhou, Li, and Liu}]{JMLR:v21:20-074}
Raffel, C.; Shazeer, N.; Roberts, A.; Lee, K.; Narang, S.; Matena, M.; Zhou,
  Y.; Li, W.; and Liu, P.~J. 2020.
\newblock Exploring the Limits of Transfer Learning with a Unified Text-to-Text
  Transformer.
\newblock \emph{Journal of Machine Learning Research}, 21(140): 1--67.

\bibitem[{Raghavan et~al.(2014)Raghavan, Fosler-Lussier, Elhadad, and
  Lai}]{raghavan-etal-2014-cross}
Raghavan, P.; Fosler-Lussier, E.; Elhadad, N.; and Lai, A.~M. 2014.
\newblock Cross-narrative Temporal Ordering of Medical Events.
\newblock In \emph{Proceedings of the 52nd Annual Meeting of the Association
  for Computational Linguistics (Volume 1: Long Papers)}, 998--1008. Baltimore,
  Maryland: Association for Computational Linguistics.

\bibitem[{Rajani et~al.(2019)Rajani, McCann, Xiong, and
  Socher}]{rajani-etal-2019-explain}
Rajani, N.~F.; McCann, B.; Xiong, C.; and Socher, R. 2019.
\newblock Explain Yourself! Leveraging Language Models for Commonsense
  Reasoning.
\newblock In \emph{Proceedings of the 57th Annual Meeting of the Association
  for Computational Linguistics}, 4932--4942. Florence, Italy: Association for
  Computational Linguistics.

\bibitem[{Ramanujan(1991)}]{ramanujan1991three}
Ramanujan, A.~K. 1991.
\newblock Three hundred Ramayanas: Five examples and three thoughts on
  translation.
\newblock \emph{Many Ramayanas: The Diversity of a Narrative Tradition in South
  Asia}, 22--49.

\bibitem[{Reichenbach(1947)}]{reichenbach1947elements}
Reichenbach, H. 1947.
\newblock Elements of symbolic logic.
\newblock \emph{London: Dover Publications (1947)}.

\bibitem[{Reichert et~al.(2010)Reichert, Kaufman, Bloxham, Chase, and
  Elhadad}]{Reichert2010CognitiveAO}
Reichert, D.; Kaufman, D.; Bloxham, B.; Chase, H.; and Elhadad, N. 2010.
\newblock Cognitive analysis of the summarization of longitudinal patient
  records.
\newblock \emph{AMIA ... Annual Symposium proceedings. AMIA Symposium}, 2010:
  667--71.

\bibitem[{Schank(1972)}]{schank1972conceptual}
Schank, R.~C. 1972.
\newblock Conceptual dependency: {A} theory of natural language understanding.
\newblock \emph{Cognitive psychology}, 3(4): 552--631.

\bibitem[{Schleppegrell(2001)}]{schleppegrell2001linguistic}
Schleppegrell, M.~J. 2001.
\newblock Linguistic features of the language of schooling.
\newblock \emph{Linguistics and education}, 12(4): 431--459.

\bibitem[{See et~al.(2019)See, Pappu, Saxena, Yerukola, and
  Manning}]{see2019massively}
See, A.; Pappu, A.; Saxena, R.; Yerukola, A.; and Manning, C.~D. 2019.
\newblock Do Massively Pretrained Language Models Make Better Storytellers?
\newblock In \emph{Proceedings of the 23rd Conference on Computational Natural
  Language Learning (CoNLL)}, 843--861. Hong Kong, China: Association for
  Computational Linguistics.

\bibitem[{Wang et~al.(2021)Wang, Liu, Zhu, Shou, Gong, Xu, and
  Zeng}]{wang2021retrieval}
Wang, H.; Liu, Y.; Zhu, C.; Shou, L.; Gong, M.; Xu, Y.; and Zeng, M. 2021.
\newblock Retrieval Enhanced Model for Commonsense Generation.
\newblock In \emph{Findings of the Association for Computational Linguistics:
  ACL-IJCNLP 2021}, 3056--3062. Online: Association for Computational
  Linguistics.

\bibitem[{Wiegreffe and Marasovic(2021)}]{wiegreffe2021teach}
Wiegreffe, S.; and Marasovic, A. 2021.
\newblock Teach Me to Explain: A Review of Datasets for Explainable Natural
  Language Processing.
\newblock In \emph{Thirty-fifth Conference on Neural Information Processing
  Systems Datasets and Benchmarks Track (Round 1)}.

\bibitem[{Wingate(2012)}]{wingate2012argument}
Wingate, U. 2012.
\newblock ‘Argument!’helping students understand what essay writing is
  about.
\newblock \emph{Journal of English for academic purposes}, 11(2): 145--154.

\bibitem[{Zhang et~al.(2019)Zhang, Kishore, Wu, Weinberger, and
  Artzi}]{zhang2019bertscore}
Zhang, T.; Kishore, V.; Wu, F.; Weinberger, K.~Q.; and Artzi, Y. 2019.
\newblock BERTScore: Evaluating Text Generation with BERT.
\newblock In \emph{International Conference on Learning Representations}.

\end{thebibliography}

\newpage
\appendix
\section*{Appendices}


\section{Discussion: NAREOR as a Hard Instance of Controllable Generation}
\label{subsec:controllable_justification}
In a simple generation task, models have to learn to generate a natural language output $o$ given some input $i$. Controllable generation tasks \cite{peng2018towards,elder2018e2e,keskar2019ctrl} differ in that models must generate $o$ given $i$ and an additional ``control variable" or goal $g$. $g$ typically is a desired property of $o$ not already determined by $i$, such as sentiment, politeness, and so forth. NAREOR is an instance of controllable generation at the story level, where $i$ is the original story $S$, $g$ is the target order $\pi_{i'}$, and $o$ is the reordered story $S'$. NAREOR is non-trivial, due to:
\begin{enumerate}
\item \textbf{Complex Goal, Challenging Invariant}: $g=\pi_{i'}$ can take on many distinct values (n!-1). Typical goal variables for controllable generation are either sparse sets (e.g. keywords) or discrete variables with few possible values (e.g. sentiment/style). Handling a many-valued variable is harder as outputs must be varied over a larger scale with variation in $g$ while maintaining faithfulness to $i$. The invariant $o_{/g}$ denotes input faithfulness, which should have the same value across outputs $o$ even for different $g$. For NAREOR, $o_{/g}$ represents whether the output plot matches the input plot (i.e. plot-preservation). This is difficult to maintain as it requires a strong understanding of story event order, characters, interactions between them, and ensuring these factors all stay consistent.
\item \textbf{Extra-Sentential, Discourse Sensitive}: Existing controllable generation tasks, such as data-to-text generation with keyword goals \cite{elder2018e2e}, generate sentence-level outputs. NAREOR requires generating a full story. As described in \S\ref{sec:dataset_analysis}, performing well on NAREOR entails learning several types of discourse dependencies such as ellipsis, coreferences, time expressions, etc. Though there are multiple-sentence generation tasks, they do not usually require a model specifically good at discourse dependencies, nor assess these aspects directly.
\end{enumerate}

\section{NAREORC Annotation Details}
\label{sec:appendix_human}
We collected NAREORC using AMT over 15 days, in a manner consistent with terms of use of any sources and intellectual property and privacy rights of AMT crowd workers. Almost all source content for NAREORC was based on ROCStories \cite{mostafazadeh-etal-2016-generating}, an already publicly available and widely used dataset in NLP research.

\begin{figure*}[!ht]
\begin{subfigure}{.97\textwidth}
    \centering
    \includegraphics[width=0.99\textwidth]{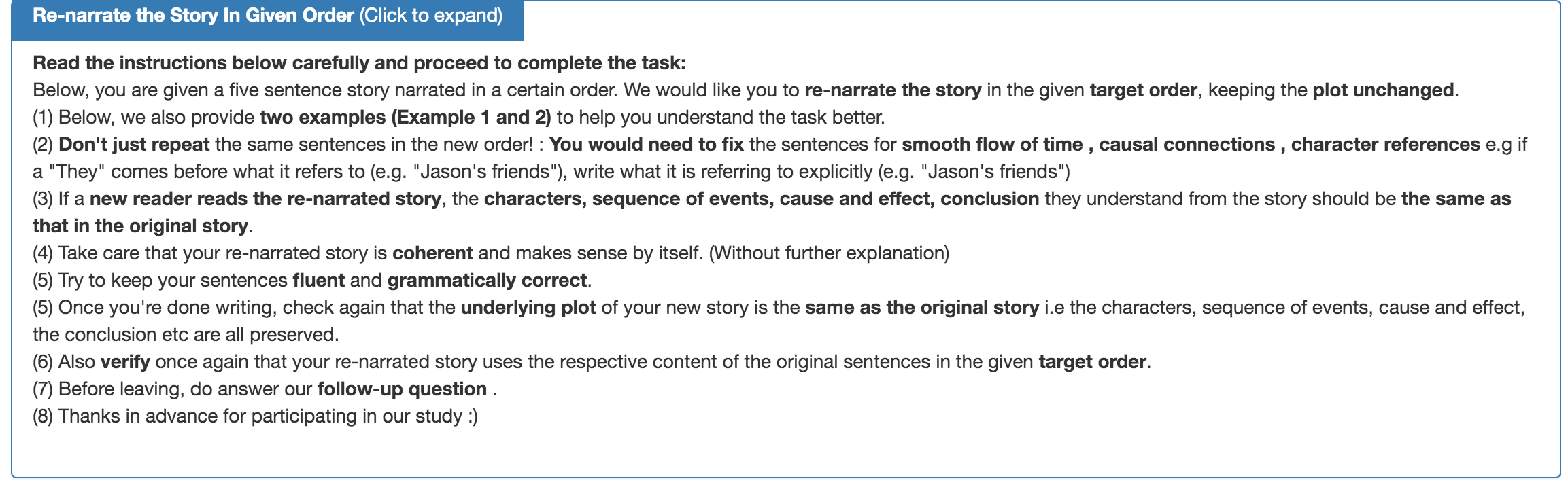}
    \caption{}
    \label{fig:amt_writing_instructions_graph}
\end{subfigure} \\
\begin{subfigure}{.97\textwidth}
    \centering
    \includegraphics[width=0.99\textwidth]{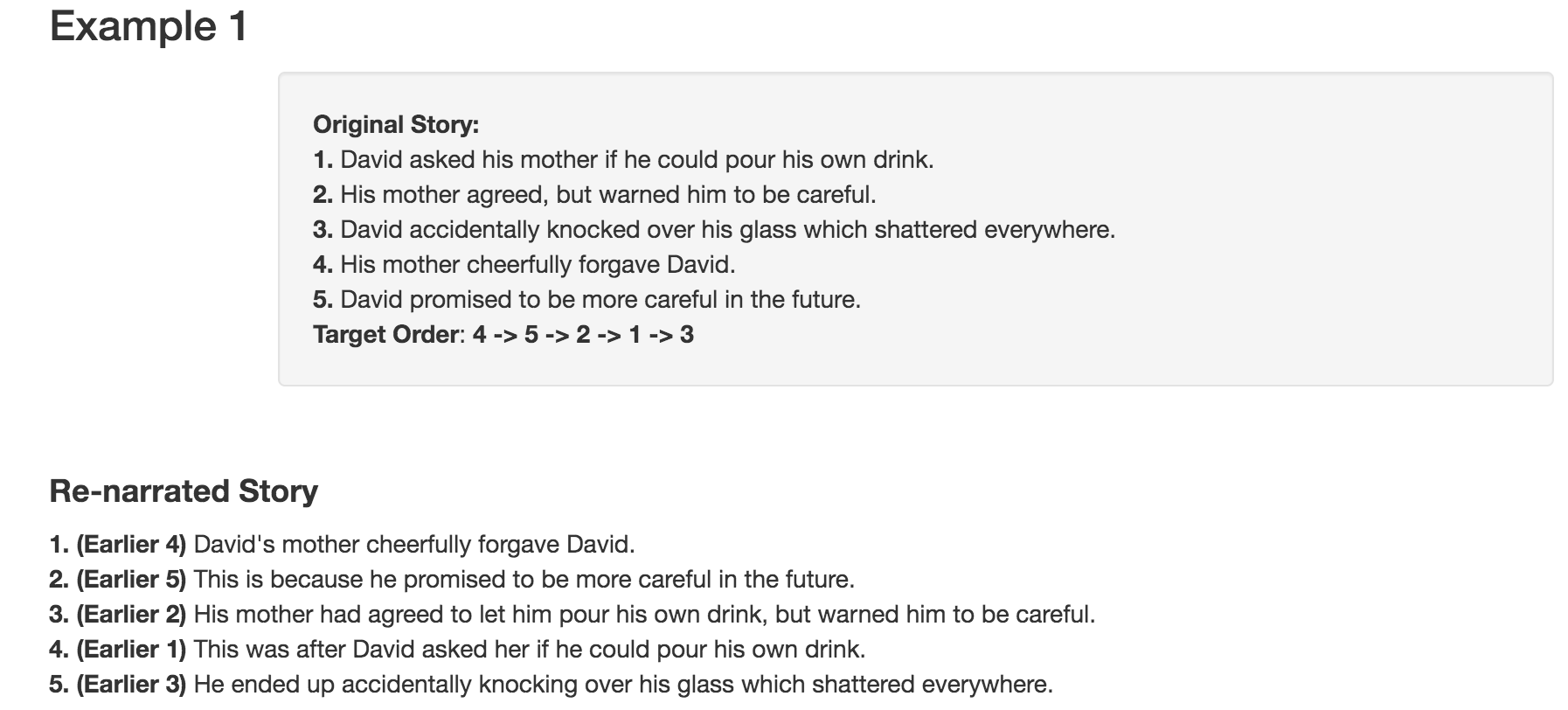}
    \caption{}
    \label{fig:amt_writing_examples_graph}
\end{subfigure} \\
\begin{subfigure}{.97\textwidth}
    \centering
    \includegraphics[width=0.99\textwidth]{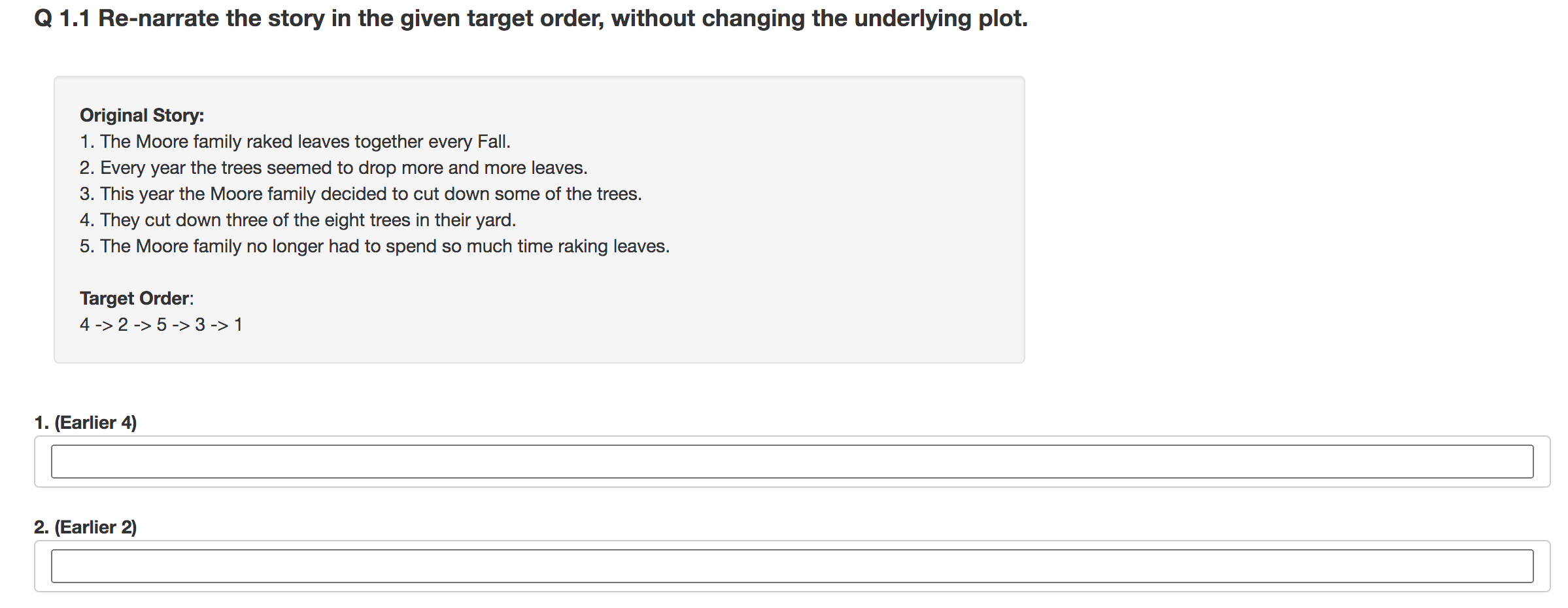}
    \caption{}
    \label{fig:amt_writing_questions_graph}
\end{subfigure}
\caption{\small Snapshots of a) instructions seen by annotator before writing, b) one of the examples provided, and c) part of the page the annotator interacts with while rewriting.\label{fig:overall_amt_writing_graph}}
\end{figure*}

We restricted annotators to be from Anglophone countries with a prior approval of $>97\%$. 
A pool of 98 different annotators participated. We manually verified each submitted story, rejecting and republishing it if found unsatisfactory. 
The hardness distribution was \textit{VeryEasy} (5.90\%), \textit{Easy} (19.03\%), \textit{Moderate} (38.73\%), \textit{Hard} (31.29\%), \textit{VeryHard} (5.03\%). 
The specific instructions, one of the examples provided, and a snippet of the question page can be seen in Figure \ref{fig:overall_amt_writing_graph}.

Crowd workers were fairly compensated: \$1.5 per rewritten story, for a roughly 5-7 min task. This is at least 1.5-2 times the minimum wage in the U.S.A. of \$7.25 per hour (\$0.725 per 6 minutes). We neither solicit, record, request, or predict any personal information pertaining to the AMT crowd workers during the annotation process.

\section{Further Model Finetuning and Generation Details}
\label{sec:appendix_finetuned}

For GPT-2 models, we use the base version with 117M parameters. We use finetuning and generation seeds of 42, and finetune with a batch size of two for the 1st stage and one for the 2nd stage. GPT2-d-1S reaches lowest validation perplexity (val-PPL) of $5.733$ after $4$ epochs, and GPT2-d-2S reaches lowest val-PPL of $7.384$ after $7$ further epochs, using LRs of 5e-5 and 5e-6, respectively. GPT2-r-1S reaches lowest val-PPL of $2.576$ after $4$ epochs, and GPT2-r-2S reaches lowest val-PPL of $4.486$ after $5$ further epochs, using an LR of 5e-6 for both.

For BART and T5, we use the base versions with 139M and 220M parameters, respectively. We use a finetuning seed of 42, and set the minimum decoder length to one token. For the NAR-d models, we use a batch size of 32, maximum encoder and decoder lengths of 128, and warmup steps of 1000 and 10 for the first and second stages, respectively. For the NAR-r models, we use a batch size of 16, maximum encoder and decoder lengths of 256, and warmup steps of 2000 and 20 for the first and second stages, respectively.

For BART-d-1S, we find lowest validation loss (val-loss) of 0.1573 after 7 epochs using LR of 3e-05 ($\approx$3hrs of train-time). For BART-d-2S, we find lowest val-loss of 0.8828 after 17 further epochs using LR of 1e-05 ($\approx$30min of train-time). For BART-r-1S, we find lowest val-loss of 0.1177 after 6 epochs using LR of 1e-06 ($\approx$4hrs of train-time). For BART-r-2S, we find lowest val-loss of 0.8949 after 9 further epochs using LR of 1e-05 ($\approx$20min of train-time). 

For T5-d-1S, we find lowest val-loss of 0.1947 after 2 epochs using an LR of 1e-04 ($\approx$1hr of train-time). For T5-d-2S, we find lowest val-loss of 0.8610 after 3 further epochs using an LR of 1e-04 ($\approx$10min of train-time). For T5-r-1S, we find lowest val-loss of 0.092 after 2 epochs using an LR of 5e-06 ($\approx$1hr of train-time). For T5-r-2S, we find lowest val-loss of 0.8619 after 2 further epochs using an LR of 1e-04 ($\approx$10mins of train-time). 

Training was done using single RTX 2080 Ti and Titan Xp GPUs, and Google Colab instances which alternately used a single V100, P100, or Tesla T4 GPU. Hyperparameters such as the learning rate were determined by trying a range of values (for LR, from 5e-8 to 5e-4), and finding ones which led to good convergence behavior (e.g. validation loss or PPL decreases at a decently steady rate and reaches min. after a reasonable number of epochs).

\begin{figure*}[!ht]
\begin{subfigure}{.97\textwidth}
    \centering
    \includegraphics[width=0.99\textwidth]{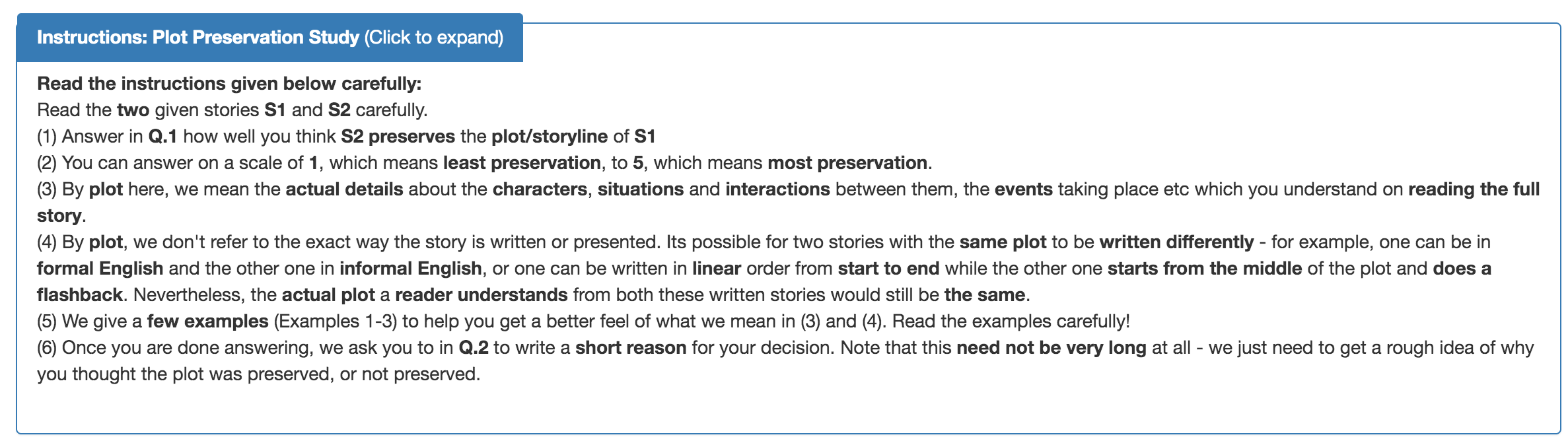}
    \caption{}
    \label{fig:amt_preserve_instructions_graph}
\end{subfigure} \\
\begin{subfigure}{.97\textwidth}
    \centering
    \includegraphics[width=0.99\textwidth]{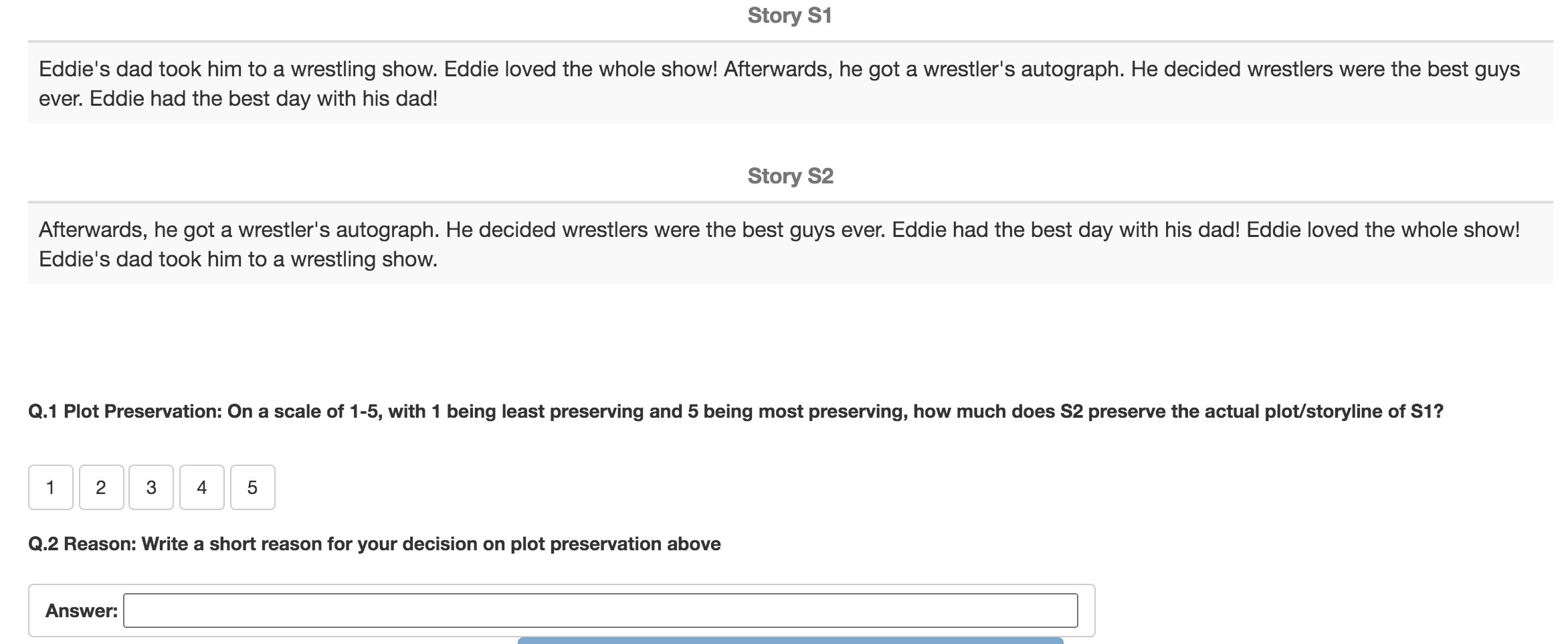}
    \caption{}
    \label{fig:amt_preserve_questions_graph}
\end{subfigure}
\caption{\small Snapshots of Plot-Preservation study: a) instructions seen by annotator and b) part of the page annotators interact with while answering the question.\label{fig:overall_amt_preserve_graph}}
\end{figure*}

\section{Human Evaluation Study Details}
\label{sec:appendix_human_eval}

The five human evaluation metrics were split into three AMT studies - one for Fluency, Coherence, and Logic, one for Plot Preservation, and one for Interestingness. Annotators were from Anglophone countries with $>97$\% approval rate. Evaluating Interestingness and Plot-pres requires more effort and reading two stories rather than one. To better verify whether annotators completed Plot-pres thoughtfully, we requested them to fill in a \textit{Reason} for their Plot-pres score. Specific instructions and a question snippet can be seen in Figure \ref{fig:overall_amt_preserve_graph}.

Crowd workers were fairly compensated: \$0.50 per fluency, coherence, and logic evaluation, \$0.50 per Plot-pres evaluation, and \$0.50 per interestingness evaluation. These are all roughly 2 minute tasks, and hence compensation is at least 1.5-2 times the minimum wage in the U.S.A. of \$7.25 per hour (or \$0.25 per 2 minutes). We neither solicit, record, request, or predict any personal information pertaining to the AMT crowd workers during the evaluation studies.

\section{Further Qualitative Examples of Rewritten Text}
\label{sec:appendix_qualitative_examples}
See Tables \ref{tab:qualitative_appendix_1} and \ref{tab:qualitative_appendix_2} for further qualitative examples using various approaches for testSup stories.

\begin{table*}[ht!]
\centering
\small
\addtolength{\tabcolsep}{-4pt}
\begin{tabular}{|p{2.3cm}|p{15.1cm}|}
\hline 
\textbf{Method} & \textbf{Text} \\ \hline 
Original story & {Amy was playing Clue with her friends. She listened to the evidence. Then she reviewed her suspects. Amy decided on her suspect. She won the game by guessing correctly.} \{2,5,4,3,1\} \\ \hline
Human rewriting & {{Amy listened closely to the evidence while playing Clue. It paid off when she won the game. She had decided carefully on her suspect. Reviewing them one by one. Her careful decision allowed her to beat the friends she was playing with.}} \\ \hline 
GPT2-d-2S & {{Amy listened to the evidence. She won the game by guessing correctly. She had the idea that she had chosen a suspect. Her latest research revealed a game she had played with her friends. When it came to games, she was actually quite pleased with her choice of suspect.}} \\ \hline 
GPT2-r-2S & {{Amy listened to the evidence. She was able to win the game by guessing correctly. She decided on her suspect. She had reviewed her suspects' evidence. The evidence of her team was published in a newspaper.}} \\ \hline
BART-d-2S & {{Amy listened to the evidence. She won the game by guessing correctly. She decided on her suspect. Then she reviewed her suspects. This was while Amy was playing Clue with her friends.}} \\ \hline 
BART-r-2S & {{Amy listened to the evidence. She won the game by guessing correctly. She decided on her suspect. Then she reviewed her suspects. She was playing Clue with her friends.}} \\ \hline 
T5-d-2S & {{Amy listened to the evidence. She won the game by guessing correctly. She had decided on a suspect. Then she reviewed her suspects. Amy was playing Clue with her friends.}} \\ \hline 
T5-r-2S & {{Amy listened to the evidence. She won the game by guessing correctly. She decided on her suspect. Then she reviewed her suspects. She was playing Clue with her friends.}} \\ \Xhline{3\arrayrulewidth}
Original story & {My Dad tried to buy a car yesterday. He went to the local dealership. He had some ideas in mind. But the dealer kept pushing other options. Dad got what he wanted, though.} \{5,3,2,1,4\} \\ \hline 
Human rewriting & {{My Dad ultimately got what he wanted from the dealership. My dad had ideas in his mind for buying a car. With these ideas, My Dad had gone to the local dealership. My Dad tried to purchase a car while at the local dealership. The dealer at the dealership, however, tried pushing other options to My Dad.}} \\ \hline 
GPT2-d-2S & {{My Dad got what he wanted for Christmas. He had some ideas in mind. He went to the local dealership. He ended up trying to buy a car yesterday. But the dealer kept pushing other options.}} \\ \hline 
BART-d-2S & {{My Dad got what he wanted, though. He had some ideas in mind. He went to the local dealership. He tried to buy a car yesterday. But the dealer kept pushing other options.}} \\ \hline 
T5-d-2S & {{My Dad got what he wanted, though. He had some ideas in mind. He went to the local dealership. He had tried to buy a car yesterday. The dealer kept pushing other options.}} \\ \hline
T5-r-2S & {{My Dad got what he wanted, though. He had some ideas in mind. He went to the local dealership. He had tried to buy a car yesterday. But the dealer kept pushing other options.}} \\ \Xhline{3\arrayrulewidth}
Original story & {Matt saw a commercial for new shoes. And they had springy looking cushions in the sole. Matt thought it would allow him to jump higher. So he begged his mom to get the shoes for him. Matt's mom eventually got him the shoes.} \{5,1,3,4,2\} \\ \hline
GPT2-d-2S & {{Matt's mom eventually got him the shoes. Matt had seen a commercial for new shoes. Matt thought it would allow him to jump higher. He was so proud that he begged his mom to get the shoes for him. The shoes looked springy looking cushions in the shoes.}} \\ \hline 
GPT2-r-2S & {{Matt thought that springy cushions were great for jumping higher. He saw a commercial for new shoes. Matt thought it would allow him to jump higher. Matt begged his mom to get the new shoes for him. The cushions had long suede feet and had sat on the back of Matt's hip.}} \\ \hline 
BART-d-2S & {{Matt's mom eventually got him the shoes. Matt had seen a commercial for new shoes. He thought it would allow him to jump higher. So he begged his mom to get the shoes for him. The shoes had springy looking cushions in the sole.}} \\ \hline 
T5-d-2S & {{Matt's mom eventually got him the shoes. Matt had seen a commercial for new shoes. Matt thought it would allow him to jump higher. He begged his mom to get the shoes for him. The shoes had springy looking cushions in the sole.}} \\ \hline
\end{tabular}
\caption{\small Further qualitative examples for testSup. Target permutations are in brackets beside the original stories.}
\label{tab:qualitative_appendix_1}
\end{table*}

\begin{table*}[ht!]
\centering
\small
\addtolength{\tabcolsep}{-4pt}
\begin{tabular}{|p{2.3cm}|p{15.1cm}|}
\hline 
\textbf{Method} & \textbf{Text} \\ \hline 
Original story & {Danny bought a new shirt. He wore it to school the next day. As he was eating lunch, he spilled mustard on it. He scrubbed it off very hard. The mustard came out and no one was the wiser.} \{5,2,4,1,3\} \\ \hline 
Human rewriting & {{Danny was able to scrub all the mustard off his shirt before anyone noticed. It was the first time he had worn this shirt to school. The mustard didn't come off easy. The shirt was brand new. Naturally he spilled mustard on it the very first time he wore it!}} \\ \hline 
GPT2-d-2S & {{The mustard came out and no one was the wiser. Danny wore his shirt to school the next day. He scrubbed it off very hard. Danny bought a new shirt. As he was eating lunch, he spilled mustard on it.}} \\ \hline 
GPT2-r-2S & {{The mustard came out and no one was the wiser. Danny wore the shirt to school the next day. He scrubbed it off very hard. Danny bought the new shirt. As he was eating lunch, he spilled mustard on the shirt.}} \\ \hline 
BART-d-2S & {{Danny's mustard came out and no one was the wiser. He wore it to school the next day. He scrubbed it off very hard. Danny bought a new shirt. While eating lunch, he spilled mustard on it.}} \\ \hline 
BART-r-2S & {{Danny's shirt came out and no one was the wiser. He wore it to school the next day. He scrubbed it off very hard. He had bought a new shirt. As he was eating lunch, he spilled mustard on his shirt.}} \\ \hline 
T5-d-2S & {{The mustard came out and no one was the wiser. Danny wore it to school the next day. He had scrubbed it off very hard. He had bought a new shirt. As he was eating lunch, he spilled mustard on it.}} \\ \hline
T5-r-2S & {{Danny's mustard came out and no one was the wiser. He wore it to school the next day. He scrubbed it off very hard. He had bought a new shirt. As he was eating lunch, he spilled mustard on it.}} \\ \Xhline{3\arrayrulewidth}
Original story & {Agatha loved to bake cookies. One day she put a batch in to bake. However, after twenty minutes, there was no scent of cookies baking. She went to see what had happened. Agatha found the oven was turned off.} \{5,1,4,3,2\} \\ \hline 
GPT2-d-2S & {{Agatha found the oven was turned off. Agatha loved to bake cookies. She went to see what had happened. However, after twenty minutes, there was no scent of cookies in her oven. It was one day after she had put the batch in to bake cookies.}} \\ \hline 
GPT2-r-2S & {{Eventually, Agatha found that the oven was turned off and turned off in the bakery. This was a mistake which she had made when she baked a cookie. This happened when she put a batch of cookies in to bake. The bake out of the oven had not been turning out well. The process she had put in the batch wasn't finished.}} \\ \hline 
BART-d-2S & {{Agatha found out the oven was turned off. Agatha loved to bake cookies. She went to see what had happened. However, after twenty minutes, there was no scent of cookies baking. This was after she put a batch in to bake.}} \\ \hline
BART-r-2S & {{Agatha found the oven was turned off. Agatha loved to bake cookies. She went to see what had happened. After twenty minutes, there was no scent of cookies baking. She had put a batch in to bake.}} \\ \hline 
T5-d-2S & {{Agatha's oven was turned off. Agatha loved to bake cookies. She went to see what had happened. However, after twenty minutes, there was no scent of cookies baking. She had put a batch in the oven to bake.}} \\ \hline
T5-r-2S & {{Agatha found the oven was turned off. Agatha had always loved to bake cookies. Agatha went to see what had happened. However, after twenty minutes, there was no scent of cookies baking. One day Agatha put a batch in to bake.}} \\ \hline
\end{tabular}
\caption{\small Further qualitative examples for testSup. Target permutations are in brackets beside the original stories.}
\label{tab:qualitative_appendix_2}
\end{table*}

\end{document}